\documentclass[10pt,twocolumn,letterpaper]{article}

\usepackage[pagenumbers]{cvpr}

\usepackage{amsmath,amssymb}
\usepackage{booktabs}
\usepackage{multirow}
\usepackage{makecell}
\usepackage{xspace}
\usepackage{enumitem}
\usepackage{colortbl}
\usepackage{tabularx}
\usepackage{algorithm}
\usepackage{algpseudocode}
\usepackage{placeins}
\definecolor{tabhighlight}{RGB}{232,245,255}
\newcolumntype{Y}{>{\centering\arraybackslash\leavevmode}X}

\DeclareMathOperator{\softmax}{softmax}
\newcommand{\loss}{\mathcal{L}}
\newcommand{\KL}{\mathrm{KL}}
\newcommand{\ours}{\textsc{DeTAR}\xspace}

\definecolor{cvprblue}{rgb}{0.21,0.49,0.74}
\usepackage[pagebackref,breaklinks,colorlinks,allcolors=cvprblue]{hyperref}

\title{Semantic Delta Tokens for Autoregressive Video Generation}

\author{
Liyu Jia$^{1*}$ \quad Mingyuan Zhang$^{1*\dagger}$ \quad Wang Lin$^{2}$ \quad Weijia Wu$^{3,2}$ \quad Xiaoyu He$^{1}$ \quad Hanwang Zhang$^{1}$\\
{\small $^1$Nanyang Technological University, $^2$Zhejiang University, $^3$National University of Singapore}\\
{\tt\small ji0011yu@e.ntu.edu.sg, zhangmy718@gmail.com}
}

\begin{document}
\maketitle
{\let\thefootnote\relax\footnotetext{$^*$Equal contribution.}\footnotetext{$^\dagger$Corresponding author.}}
\begin{abstract}
Autoregressive (AR) generation is essential for long videos, as each segment builds on the last.
AR diffusion models entangle motion with appearance and model physics poorly.
Agents separate them only via slow tool calls.
Prior decoupled designs train tokenizers for reconstruction and confine causality to short clips.
We decouple motion from appearance with semantic delta tokens: an AR model predicts motion as tokens, which a diffusion model renders.
Our tokens have three properties.
(1)~Semantic: predicting semantic feature evolution, not reconstructing appearance, makes them easier to learn.
(2)~Chunk-level: chunks are encoded independently, enabling chunk-wise causal generation of long videos.
(3)~Delta: encoding only inter-frame change, 32 tokens per chunk suffice, nearly 50$\times$ fewer than a 480p frame's roughly 1{,}500 SigLIP tokens.
Built on these tokens, we present \textsc{DeTAR}, which pairs the renderer with a vision-language model reasoning natively from the first frame, optionally with video or text.
Extensive experiments show that \textsc{DeTAR} captures physical dynamics, generates well beyond five-second training clips, transfers motion zero-shot, and enables fine-grained temporal control.
Project page: \url{https://liyu002.github.io/DeTAR/}.
\end{abstract}

\section{Introduction}
\label{sec:intro}

\suppressfloats[t]
\begin{figure}[t]
  \centering
  \includegraphics[width=\linewidth]{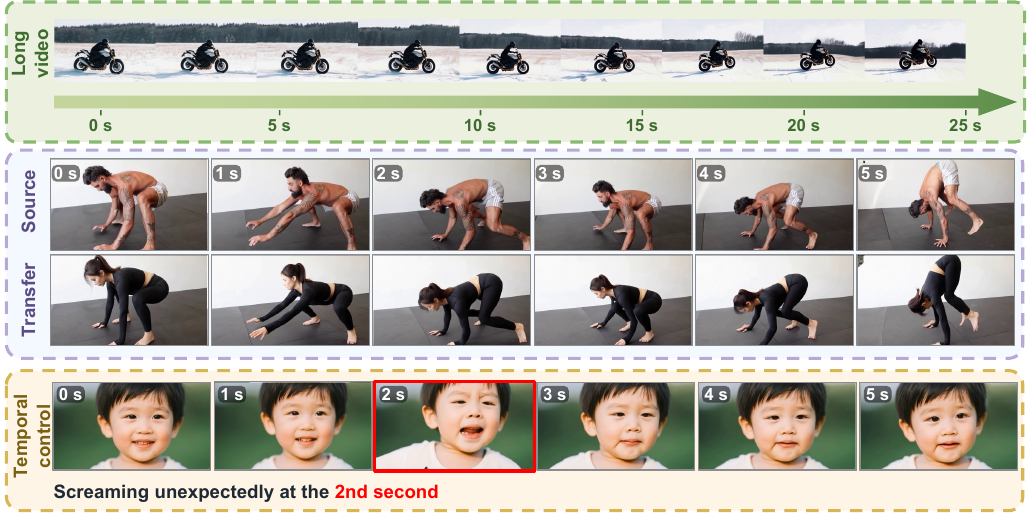}
  \caption{Semantic delta tokens decouple AR reasoning from diffusion rendering, enabling long video generation, zero-shot motion transfer, and temporal control.}
  \label{fig:teaser}
\end{figure}

Generating long videos requires each new segment to build on the last~\cite{streamingt2v,fifodiffusion,framepack,skyreelsv2,selfforcingpp,longlive} (\eg, when a ball is thrown, bounces twice, and rolls under a table, where it ends up after ten seconds depends on every bounce before it).
Autoregressive (AR) generation fits this structure, since it conditions every step on the history.
The dominant approach, AR diffusion~\cite{diffusionforcing,causvid,selfforcing,magi1,nova,pyramidflow}, denoises future frames from past ones with a single diffusion model, entangling frame evolution with appearance and modeling physics poorly.
This calls for decoupling them: an AR language model can reason causally about how frames evolve, while a diffusion model renders appearance in pixel-level detail.

However, decoupling well is challenging, because the tokens passed from the AR model to the renderer need to meet several requirements.
First, the tokens should be \mbox{semantic}, so that the AR model can reason over them as easily as over text.
In addition, the tokens should be chunk-level, so that the AR model can reason over them causally and continuously instead of being tied to short clips of a fixed length.
Moreover, the tokens should be compact, since the AR model needs to model how frames evolve rather than each frame itself, and the sequence it reasons over grows with the length of the video.

Several recent works attempt to decouple reasoning from rendering.
Agents~\cite{videodirectorgpt,freebloom,vlogger,mora,vgot,movieagent} let a language model plan each clip in text and call a diffusion model as a tool to render it.
Besides being slow, they pass only text and at most one image from one clip to the next, so motion states such as velocity, acceleration, and trajectory are difficult to convey.
Another line of work replaces text with learned tokens that a diffusion model renders, but these tokens each miss one of the requirements: they are not semantic when learned by reconstructing VAE latents~\cite{videoworld2,phizero}, not chunk-level when tied to keyframes~\cite{landiff,vtok}, and not compact when describing a full state for every frame~\cite{planx,bividgen}.

To address the limitations of these methods and decouple reasoning from rendering properly, we propose \emph{semantic delta tokens}: an AR model predicts these tokens chunk by chunk, and a diffusion model renders them into video.
Our tokenizer builds the tokens in the semantic feature space of SigLIP2~\cite{siglip2}.
Its encoder takes two consecutive frames one third of a second apart and encodes only how the second frame evolves from the first.
The decoder then takes the first frame and these delta tokens and infers the second frame.
These delta tokens are therefore compact: 32 tokens suffice for each pair of frames, nearly 50$\times$ fewer than the roughly 1{,}500 SigLIP tokens of a 480p frame.
However, learning them by regression rewards reproducing every detail of the target, so the semantics in the delta tokens can be overwhelmed by noise when the evolution between two frames is small.
We therefore learn the tokens by predicting distributions over the frozen X-Omni~\cite{xomni} codebook instead of regressing features, so that the tokens follow changes in semantics rather than noise in the features.
The two frames form a chunk that is encoded independently, so generation can continue chunk by chunk beyond the training horizon.

Built on these tokens, we present \ours (Semantic \textbf{De}lta \textbf{T}okens for \textbf{A}uto\textbf{R}egressive Video Generation), which pairs a VLM as the AR model with a separately trained diffusion renderer.
As shown in Figure~\ref{fig:teaser}, \ours generates long videos, transfers motion zero-shot, and follows timing instructions.
Since the chunks are encoded independently, the AR model must learn the dynamics between chunks from abundant data, so we pretrain it on unlabeled videos in a video-to-video (V2V) manner.
Existing V2V pretraining~\cite{cosmos} relies on dense, non-semantic tokens that limit it to 33 frames, whereas our semantic tokens fit clips of up to 121 frames in only 480 tokens, and this pretraining unlocks the reasoning of the AR model more effectively than training directly for text-image-to-video (TI2V) generation.
Trajectory distillation on its own rollouts further improves the AR model.
Extensive experiments show that our approach captures physical dynamics, generates well beyond its five-second training clips, transfers motion zero-shot, and enables fine-grained temporal control.
Our contributions are three-fold:
\begin{itemize}[leftmargin=*,nosep]
\item We propose semantic delta tokens, learned by predicting distributions over a frozen semantic codebook, to properly decouple the AR reasoner from the renderer.
\item We present \ours, which decouples a VLM from a diffusion renderer. The VLM is pretrained on unlabeled videos in a V2V manner and post-trained with trajectory distillation, which provides a better starting point for the downstream TI2V task.
\item We achieve SOTA results on V2V generation, physics-aware TI2V generation, long video generation, and temporal control, and competitive results on general video benchmarks and video editing as a zero-shot method.
\end{itemize}

\section{Related Work}
\label{sec:related}

\begin{figure*}[t]
  \centering
  \includegraphics[width=\linewidth]{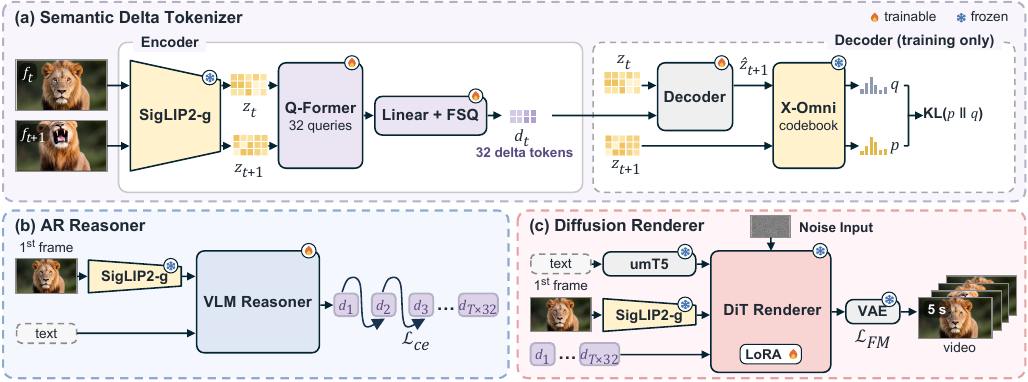}
  \caption{\textbf{Overview of our approach.}
  (a) The tokenizer encodes how $f_{t+1}$ evolves from $f_t$ into 32 semantic delta tokens, supervised by predicting distributions over the frozen X-Omni codebook, and its decoder is used only for training.
  (b) A VLM predicts the delta tokens chunk by chunk from the first frame and optional text.
  (c) A separate diffusion renderer uses the tokens as conditioning signals.}
  \label{fig:overview}
\end{figure*}

Long videos are commonly generated by AR diffusion~\cite{diffusionforcing,historyguidance,causvid,selfforcing,magi1,nova,pyramidflow,framepack,streamingt2v,fifodiffusion,skyreelsv2,selfforcingpp,longlive,adaworld,videomar}, where a single diffusion model models frame evolution and appearance jointly.
Agents~\cite{videodirectorgpt,freebloom,vlogger,mora,dreamfactory,vgot,genmac,movieagent} let a language model call a diffusion model as a tool for each clip, but they rely on slow tool calls, and their text interface can hardly carry the motion state across clips.
Other methods plan in learned visual representations instead.
Bernini~\cite{bernini} lets an MLLM predict ViT features of the target video, and PAN~\cite{pan} predicts latent states with an LLM-based backbone, but neither plans by next-token prediction.
Plan-X~\cite{planx} predicts TA-Tok tokens quantized from SigLIP2 features, BiVidGen~\cite{bividgen} predicts tokens from the Qwen3-VL vision encoder, and ARLON~\cite{arlon} predicts coarse latent tokens for a diffusion model to refine, but all of them predict a full state for every frame or keyframe.
LanDiff~\cite{landiff} and VTok~\cite{vtok} encode change only relative to keyframes, which binds their tokens to the keyframe and to the length of the clip.
LanDiff further re-predicts a 330-token keyframe for every group of frames.
VideoWorld~2~\cite{videoworld2} learns per-step change codes and PhiZero~\cite{phizero} learns clip-level tokens, both by reconstructing VAE latents through a diffusion decoder, and the tokens of PhiZero model temporal order poorly.
DeltaWorld~\cite{deltaworld} encodes the difference between consecutive features of a vision foundation model into a single continuous token for efficient dense forecasting, and DeltaWAM~\cite{deltawam} builds on it for real-time robotic manipulation.
Such continuous tokens cannot be predicted by a VLM through next-token prediction.
Latent action models~\cite{genie,lapa,moto,univla} learn discrete inter-frame tokens by reconstructing future frames or DINOv2 features for short-horizon control rather than long video rendering.
Cosmos~\cite{cosmos} pretrains an AR model on unlabeled videos with dense, non-semantic tokens, which limit it to only 33 frames.
In contrast, our semantic and compact delta tokens let the AR model pretrain on clips of up to 121 frames.

\section{Semantic Delta Tokens}
\label{sec:tokens}

To decouple AR reasoning from diffusion rendering properly, we need a tokenizer whose tokens are semantic, chunk-level, and compact, so that an AR model can reason over them chunk by chunk and a diffusion model can render them into video.
We sample keyframes every 8 frames from a 24\,fps video, so two consecutive keyframes $f_t$ and $f_{t+1}$ are one third of a second apart and form a chunk.
For each chunk, the tokenizer encodes only how $f_{t+1}$ evolves from $f_t$ into 32 discrete delta tokens, working entirely in the semantic feature space of SigLIP2~\cite{siglip2} (\cref{fig:overview}a).
It consists of three parts.
A frozen encoder $E$ maps each frame to semantic features $z_t = E(f_t)$.
A quantizer $Q$ then compresses the change from $z_t$ to $z_{t+1}$ into the delta tokens $d_t = Q(z_t, z_{t+1})$, which form a tight bottleneck.
Finally, a decoder $D$ combines $d_t$ with $z_t$ to infer the features of the second frame, $\hat z_{t+1} = D(z_t, d_t)$.
Instead of reconstructing $z_{t+1}$ itself, the tokenizer is trained to predict the distribution of the second frame over a frozen visual codebook.

\noindent\textbf{Encoder.}
The encoder is the SigLIP2-g/16 vision encoder that X-Omni~\cite{xomni} uses for its visual tokens, and we keep it frozen.
We take its 1536-d features after the last of its 40 blocks and before the final normalization.
We standardize these features per channel with the mean and standard deviation estimated on training frames to obtain $z_t$.

\noindent\textbf{Quantizer.}
A Q-Former~\cite{blip2} with 32 learnable queries cross-attends to $z_t$ and $z_{t+1}$ and outputs 32 continuous vectors.
Its keys carry fixed 2D sinusoidal position embeddings and learnable frame-slot embeddings that distinguish $f_t$ from $f_{t+1}$, and its values carry only the frame-slot embeddings.
Two linear layers project each vector from 1536 to 48 and then to 6 dimensions, and FSQ~\cite{fsq} quantizes the 6 dimensions to $8,5,5,5,5,5$ levels, respectively, which gives a vocabulary of $8\!\times\!5^5 = 25{,}000$ delta tokens.
A chunk thus needs only 32 tokens, nearly 50$\times$ fewer than the 1{,}404--1{,}680 SigLIP patch features of a 480p frame across our aspect-ratio buckets.
Each chunk is quantized independently with no attention across chunks, so $d_t$ describes only the step from $f_t$ to $f_{t+1}$, and a five-second clip yields a causal sequence of 15 chunks and 480 tokens.

\noindent\textbf{Decoder.}
The decoder consists of 24 transformer blocks initialized from the last 24 blocks of SigLIP2-g.
We add the SigLIP position embeddings to $z_t$, embed the 32 tokens of $d_t$ with an MLP and add learnable position embeddings to them, and concatenate the two parts into the input sequence.
The outputs at the positions of $z_t$ give the predicted features $\hat z_{t+1}$ of $f_{t+1}$.
Because the decoder already sees $z_t$, the tokens only need to encode how $f_{t+1}$ evolves from $f_t$, which makes them delta tokens.
The decoder is discarded after training.

\noindent\textbf{Training.}
Instead of regressing $\hat z_{t+1}$ onto $z_{t+1}$, we compare the two through the frozen visual codebook $\mathcal{C}$ of X-Omni ($|\mathcal{C}| = 16{,}384$).
With $\phi$ denoting the inverse standardization followed by X-Omni's frozen projection and $\ell_2$ normalization, the predicted and target distributions at each patch are
\begin{align}
  q &= \softmax\bigl(s \cdot \cos(\phi(\hat z_{t+1}),\, \mathcal{C})\bigr), \label{eq:pred}\\
  p &= \softmax\bigl(\cos(\phi(z_{t+1}),\, \mathcal{C})/\tau\bigr), \label{eq:target}
\end{align}
where $\tau$ is a fixed temperature.
Since cosine similarities lie in $[-1,1]$, a softmax over them would be nearly uniform across the 16{,}384 entries, so both distributions need to be sharpened.
The target uses the fixed $1/\tau$, while $s$ is a learnable scale initialized to $1/\tau$ and clamped to $[10,100]$.
The total loss, averaged over patches, is
\begin{equation}
  \loss = \KL\!\bigl(p \,\|\, q\bigr) + \lambda_e\,\loss_{\text{ent}},
  \label{eq:loss}
\end{equation}
with an FSQ entropy regularizer $\loss_{\text{ent}}$ and an empirically set weight $\lambda_e\!=\!0.01$, following~\cite{phizero}.
Because the target is a distribution over semantic tokens rather than exact feature values, the tokens follow changes in semantics rather than noise in the features.

\section{\ours}
\label{sec:detar}

With the delta tokens, \ours assigns predicting how the video evolves to a pretrained VLM to leverage the reasoning ability it learned from text, and assigns rendering how the video looks to a diffusion model, which renders the tokens in pixel-level detail (\cref{fig:overview}b and c).

\begin{figure*}[t]
\centering
\includegraphics[width=\linewidth]{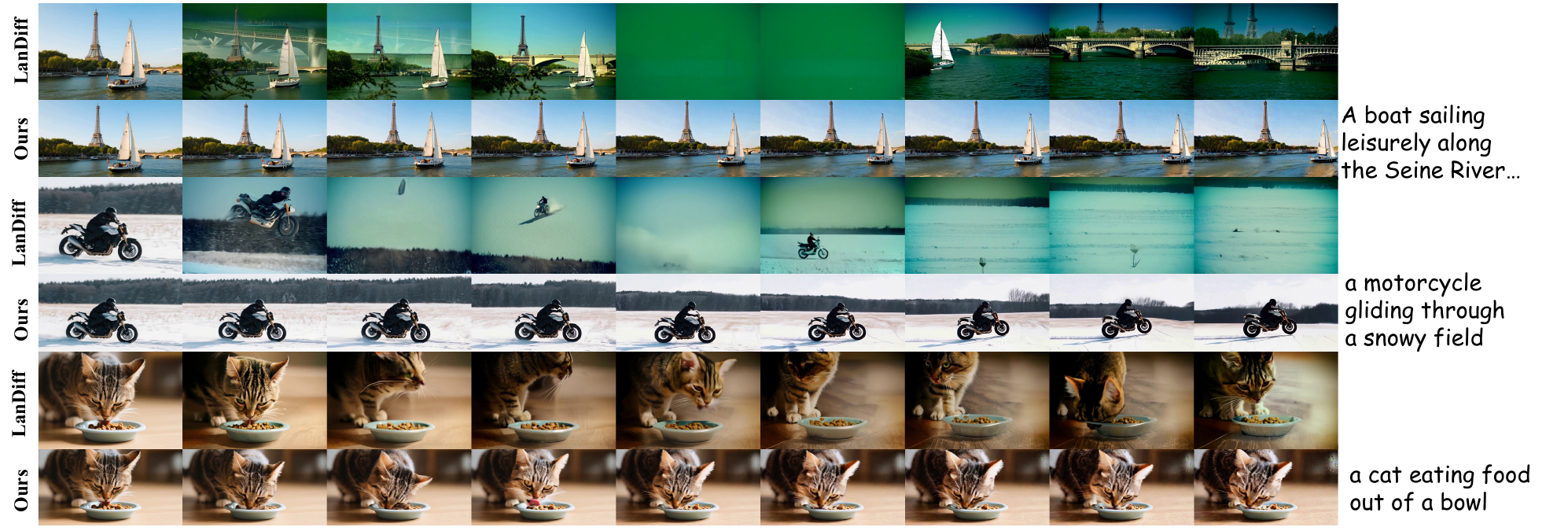}
\caption{\textbf{Long video generation on VBench-Long.} The first frames are generated by Wan2.2-TI2V-5B. LanDiff drifts in color and content over time, while \ours keeps the subject and the scene consistent over the whole video.}
\label{fig:long}
\end{figure*}

\subsection{Autoregressive Vision-Language Model}
\label{sec:ar}

We initialize the AR model from Qwen3-VL-8B~\cite{qwen3vl} and expand its vocabulary with the 25{,}000 delta tokens.
Their input and output embeddings are initialized from the MLP that embeds the tokens in the tokenizer decoder.
We further add 32 slot embeddings, initialized from the learnable position embeddings that the tokenizer decoder adds to the tokens, to mark the position of each token within its chunk.
To let it read the first frame in the same feature space as the delta tokens, we replace its vision encoder with the frozen SigLIP2-g of the tokenizer, followed by a trainable projection.
All parameters except SigLIP2-g are trained in all training stages.

\noindent\textbf{Stage 1: V2V pretraining.}
Captioned videos are costly to collect, while unlabeled videos are easy to obtain.
We therefore first pretrain the AR model on 5.7M unlabeled clips of 2 to 5 seconds, where it takes the first frame of each clip and the ground-truth tokens of its first second (96 tokens) and predicts the following tokens.
Owing to the nature of AR training, the model can then continue any clip of one second or longer.
A five-second clip takes only 480 tokens, so the model trains at a modest sequence length.

\noindent\textbf{Stage 2: V2V trajectory distillation.}
Teacher forcing trains the model on ground-truth prefixes, whereas at inference it conditions on its own predictions.
To close this gap, inspired by~\cite{opsd}, we let the pretrained model learn from its own trajectories under a teacher that sees more context.
For each unlabeled clip, let $x$ denote the later ground-truth tokens, starting from a later keyframe, and $y^*$ the earlier ground-truth tokens before them.
The student $p_S(\cdot\mid x)$, \ie, the model $p_\theta$ given only $x$, samples its own rollout $\hat y \sim p_S(\cdot \mid x)$.
The teacher $p_T(\cdot\mid x,y^*)$, a frozen copy of the pretrained model given both $x$ and $y^*$, is teacher-forced on $\hat y$.
We minimize
\begin{equation}
  \loss_{\text{TD}} = \mathbb{E}\tfrac{1}{|\hat y|}\textstyle\sum_{n}\!\KL\bigl(p_T(\cdot|x,y^*\!,\hat y_{<n})\|p_S(\cdot|x,\hat y_{<n})\bigr),
  \label{eq:td}
\end{equation}
where both distributions are computed by a softmax over the delta tokens only and gradients flow only through the student.
\Cref{alg:td} in the appendix summarizes the procedure.

\noindent\textbf{Stage 3: TI2V fine-tuning.}
Finally, we fine-tune the distilled model on 2.0M captioned five-second clips to predict the delta tokens from the first frame and the text prompt.
At inference, we sample the delta tokens with temperature 1 and top-$k$ 50.

\begin{figure*}[t]
\centering
\includegraphics[width=\linewidth]{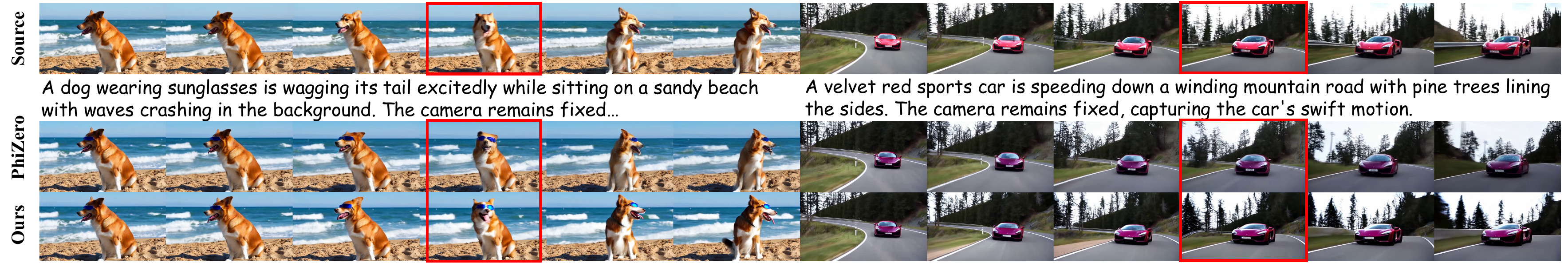}
\caption{\textbf{Zero-shot editing on FiVE.} The first frames are edited by InstructPix2Pix for all methods. Red boxes mark frames where \ours follows the motion of the source video more closely than PhiZero.}
\label{fig:five}
\end{figure*}

\begin{figure*}[t]
\centering
\includegraphics[width=\linewidth]{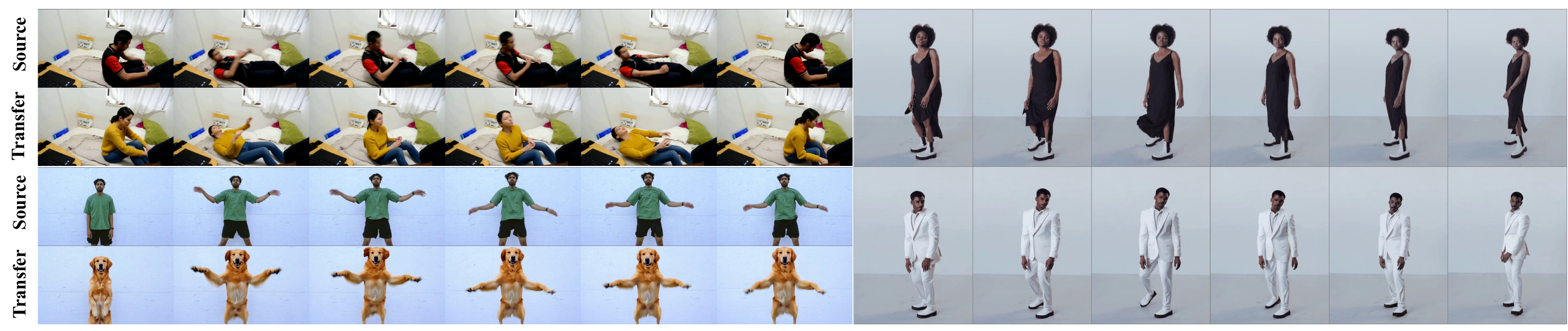}
\caption{\textbf{Motion transfer.} The first frames are edited by Qwen-Image, and \ours renders them with the delta tokens of the input video, so the motion transfers to a new subject. More examples are in \cref{sec:supp_transfer}.}
\label{fig:transfer}
\end{figure*}

\begin{figure*}[t]
\centering
\includegraphics[width=\linewidth]{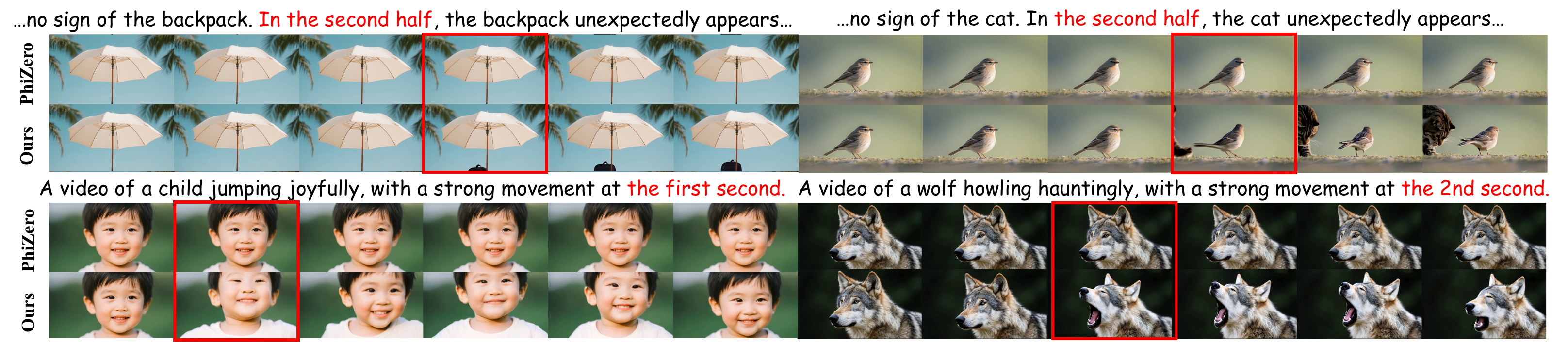}
\caption{\textbf{Temporal control on the benchmark of TempoControl.} The first frames are generated by Wan2.2-TI2V-5B. Red boxes mark the time specified by the prompt, at which the object or the movement appears only with \ours, whereas PhiZero stays almost static.}
\label{fig:tempo_vis}
\end{figure*}

\subsection{Diffusion Renderer}
\label{sec:renderer}

The renderer is Wan2.2-TI2V-5B~\cite{wan22} fine-tuned with LoRA.
Three conditions enter through cross-attention: the umT5 text embedding, the SigLIP2-g features of the first frame, and learned embeddings of the 480 delta tokens.
The delta token embeddings are initialized from the MLP that embeds the tokens in the tokenizer decoder, together with its learnable position embeddings for the 32 tokens of a chunk, and new learnable embeddings mark the 15 chunks.
Likewise, the embeddings of the SigLIP2-g features are initialized from the layer through which the tokenizer decoder reads $z_t$, and the SigLIP position embeddings are added to them as in the decoder.
These embeddings are trained together with the LoRA weights.
It is trained on the same 2.0M captioned clips and their delta tokens, independently of the AR model.
Following InstructPix2Pix~\cite{instructpix2pix}, we apply classifier-free guidance to two conditions: the visual condition $c_I$, which binds the SigLIP2-g features of the first frame with the delta tokens, and the text condition $c_T$.
During training, we randomly drop $c_T$ alone, $c_I$ alone, or both.
At inference, the velocity prediction for a noisy latent $x$ is
\begin{equation}
\begin{aligned}
  \tilde v_\theta(x, c_I, c_T) ={}& v_\theta(x, \varnothing, \varnothing) \\
  &+ s_I \bigl(v_\theta(x, c_I, \varnothing) - v_\theta(x, \varnothing, \varnothing)\bigr) \\
  &+ s_T \bigl(v_\theta(x, c_I, c_T) - v_\theta(x, c_I, \varnothing)\bigr),
\end{aligned}
\label{eq:cfg}
\end{equation}
where $s_T$ amplifies the information that the text adds beyond the first frame and the delta tokens.
When the AR model continues a video beyond what the text prompt describes (\eg, when the prompt covers only the first five seconds), the continued segment is rendered without text, with the last frame of the previous segment as its first frame and the tokens from continued AR sampling.
The renderer and the AR model thus communicate only through the delta tokens, which keeps reasoning and rendering decoupled.

\section{Experiments}
\label{sec:exp}

%==========================================================
\subsection{Experimental Settings}
\label{sec:exp_setting}
\label{sec:impl}

\noindent\textbf{Data.}
All training data come from open-source datasets.
V2V pretraining uses 5.7M unlabeled clips of up to five seconds, including 5.3M real and 0.4M simulated clips, of which about 1.08M are shorter than five seconds, and the tokenizer is trained on the same clips.
Trajectory distillation, TI2V fine-tuning, and the renderer use 2.0M captioned five-second clips with high dynamics, and trajectory distillation drops the captions.
All clips are sampled at 24\,fps, so a five-second clip has 121 frames and 15 chunks, and are resized to 480p in aspect-ratio buckets, which yield 1{,}404--1{,}680 SigLIP2 patches per frame.
\Cref{tab:data} in the appendix lists the datasets.

\begin{table}[t]
\centering
\caption{V2V generation on WorldBench. w/o.~TD: only V2V pretraining. w.~TD: trajectory distillation after V2V pretraining. The best results for each number of frames are in bold, and the best results among all methods are underlined.}
\label{tab:worldbench}
\fontsize{6.5}{7.5}\selectfont\renewcommand{\arraystretch}{1.15}
\setlength{\tabcolsep}{3.6pt}
\begin{tabular}{@{}l|c|*{5}{c}@{}}
\toprule
Model & Frames & Motion$\uparrow$ & Permanence$\uparrow$ & Scale$\uparrow$ & Support$\uparrow$ & All$\uparrow$ \\
\hline
Cosmos-1-AR-5B  & \multirow{4}{*}{33} & 0.290 & 0.402 & 0.456 & 0.545 & 0.423 \\
Cosmos-1-7B     & & 0.318 & 0.459 & 0.520 & 0.478 & 0.451 \\
\ours$_{\text{w/o.\ TD}}$  & & 0.420 & 0.677 & 0.443 & 0.586 & 0.545 \\
\textbf{\ours$_{\text{w.\ TD}}$ (Ours)} & & \underline{\textbf{0.452}} & \underline{\textbf{0.786}} & \underline{\textbf{0.535}} & \underline{\textbf{0.622}} & \underline{\textbf{0.620}} \\
\hline
Cosmos-2-2B     & \multirow{3}{*}{93} & 0.169 & 0.362 & \textbf{0.347} & 0.380 & 0.320 \\
Cosmos-2-14B    & & 0.176 & 0.255 & 0.302 & 0.339 & 0.268 \\
Cosmos-2.5-2B   & & \textbf{0.221} & \textbf{0.376} & 0.346 & \textbf{0.434} & \textbf{0.349} \\
\hline
Cosmos-1-7B     & \multirow{3}{*}{121} & 0.170 & 0.328 & 0.189 & 0.320 & 0.257 \\
\ours$_{\text{w/o.\ TD}}$  & & 0.302 & 0.494 & 0.165 & 0.487 & 0.366 \\
\textbf{\ours$_{\text{w.\ TD}}$ (Ours)} & & \textbf{0.340} & \textbf{0.755} & \textbf{0.203} & \textbf{0.532} & \textbf{0.480} \\
\bottomrule
\end{tabular}
\end{table}

\begin{table*}[t]
\centering
\caption{Physics-aware TI2V generation on Physics-IQ Verified, PhyGround, and WorldModelBench.}
\label{tab:physics}
\fontsize{6.5}{7.5}\selectfont\renewcommand{\arraystretch}{1.15}
\setlength{\tabcolsep}{1pt}
\begin{tabularx}{\linewidth}{@{}l|*{4}{Y}|l|*{3}{Y}|l|*{3}{Y}@{}}
\toprule
\multirow{2}{*}{Model} & \multicolumn{4}{c|}{\textbf{Physics-IQ Verified}} & \multirow{2}{*}{Model} & \multicolumn{3}{c|}{\textbf{PhyGround}} & \multirow{2}{*}{Model} & \multicolumn{3}{c}{\textbf{WorldModelBench}} \\
 & S-IoU$\uparrow$ & ST-IoU$\uparrow$ & WS-IoU$\uparrow$ & \mbox{IQ-Score$\uparrow$} & & Quality$\uparrow$ & Physics$\uparrow$ & Overall$\uparrow$ & & Physics$\uparrow$ & \makecell{Common\\Sense$\uparrow$} & Total$\uparrow$ \\
\hline
Wan2.2-5B~\cite{wan22}       & 24.7 & 22.6 & 13.3 & 21.2 & LTX-Video-19B~\cite{ltxvideo}  & 2.56 & 2.54 & 2.55 & Pandora~\cite{pandora}            & 4.05 & 0.95 & 6.57 \\
Sora 2~\cite{sora2}          & 37.3 & 27.0 & 26.9 & 26.5 & LTX-Video-22B~\cite{ltxvideo}  & 2.59 & 2.56 & 2.57 & OpenSora-Plan~\cite{opensoraplan} & 3.85 & 1.01 & 6.62 \\
Cosmos3-Nano~\cite{cosmos3}  & 40.4 & 22.0 & 24.6 & 29.1 & Wan2.2-5B~\cite{wan22}         & 2.67 & 2.65 & 2.66 & CogVideoX~\cite{cogvideox}        & 3.88 & 0.99 & 6.75 \\
Wan2.2-14B~\cite{wan22}      & 51.1 & 20.5 & 28.5 & 32.2 & Cosmos2.5-2B~\cite{cosmos25}   & 2.79 & 2.70 & 2.74 & Mochi~\cite{mochi}                & 4.14 & 1.26 & 7.62 \\
Hunyuan-Video~\cite{hunyuan} & 47.1 & 26.9 & 29.7 & 33.4 & OmniWeaving~\cite{omniweaving} & 2.89 & 2.78 & 2.84 & Luma~\cite{luma}                  & 4.13 & 1.57 & 7.72 \\
Grok-Video~\cite{grokvideo}  & 52.7 & 21.4 & 35.7 & 34.8 & Veo3.1~\cite{veo31}            & 3.01 & 2.85 & 2.93 & Wan2.2-5B~\cite{wan22}            & 4.51 & 1.41 & 7.98 \\
Cosmos3-Super~\cite{cosmos3} & --   & --   & --   & 39.5 & Wan2.2-14B~\cite{wan22}        & 3.00 & 2.90 & 2.95 & Runway~\cite{runway}              & 4.27 & 1.65 & 8.08 \\
PhiZero~\cite{phizero}       & \textbf{58.2} & 36.8 & 27.6 & 41.2 & PhiZero~\cite{phizero} & 2.93 & 3.01 & 2.97 & PhiZero~\cite{phizero}   & 4.88 & 1.71 & 8.19 \\
\hline
\textbf{\ours (Ours)} & 52.4 & \textbf{65.0} & \textbf{46.0} & \textbf{48.1} & \textbf{\ours (Ours)} & \textbf{3.07} & \textbf{3.12} & \textbf{3.10} & \textbf{\ours (Ours)} & \textbf{4.93} & \textbf{1.91} & \textbf{8.66} \\
\bottomrule
\end{tabularx}
\end{table*}

\begin{table}[t]
\centering
\caption{Image-to-video generation on VBench-I2V.}
\label{tab:vbench}
\fontsize{6.5}{7.5}\selectfont\renewcommand{\arraystretch}{1.15}
\setlength{\tabcolsep}{6pt}
\begin{tabular*}{\linewidth}{@{\extracolsep{\fill}}l|*{3}{c}@{}}
\toprule
Model & Total Score$\uparrow$ & I2V Score$\uparrow$ & Quality Score$\uparrow$ \\
\hline
Magi-1~\cite{magi1} & \textbf{89.28} & 96.12 & \textbf{82.44} \\
Wan2.2-TI2V-5B~\cite{wan22} & 88.80 & 96.30 & 81.30 \\
Cosmos-5B~\cite{cosmos} & 84.16 & 92.51 & 75.81 \\
Cosmos-13B~\cite{cosmos} & 84.22 & 92.60 & 75.83 \\
VideoMAR~\cite{videomar} & 84.82 & 94.02 & 75.61 \\
\hline
\textbf{\ours (Ours)} & 89.12 & \textbf{97.80} & 80.43 \\
\bottomrule
\end{tabular*}
\end{table}

\begin{table}[t]
\centering
\caption{Long video generation on VBench-Long.}
\label{tab:vbench_long}
\fontsize{6.5}{7.5}\selectfont\renewcommand{\arraystretch}{1.15}
\setlength{\tabcolsep}{1.3pt}
\begin{tabular*}{\linewidth}{@{\extracolsep{\fill}}l|*{7}{c}@{}}
\toprule
Model & {\fontsize{5}{6}\selectfont\makecell{Subject\\Consistency$\uparrow$}} & {\fontsize{5}{6}\selectfont\makecell{Background\\Consistency$\uparrow$}} & {\fontsize{5}{6}\selectfont\makecell{Motion\\Smoothness$\uparrow$}} & {\fontsize{5}{6}\selectfont\makecell{Dynamic\\Degree$\uparrow$}} & {\fontsize{5}{6}\selectfont\makecell{Aesthetic\\Quality$\uparrow$}} & {\fontsize{5}{6}\selectfont\makecell{Imaging\\Quality$\uparrow$}} & {\fontsize{5}{6}\selectfont\makecell{Overall\\Consistency$\uparrow$}} \\
\hline
FreeNoise & 96.59 & 97.48 & 98.36 & 17.44 & 47.39 & 63.88 & 25.78 \\
StreamingT2V~\cite{streamingt2v} & 87.31 & 94.64 & 93.83 & \textbf{85.64} & 44.57 & 53.64 & 23.65 \\
OpenSora-V1.2 & 96.30 & 97.39 & 98.94 & 44.79 & 56.68 & 51.64 & 26.36 \\
ARLON~\cite{arlon} & 97.11 & 97.56 & 98.50 & 50.42 & 56.85 & 53.85 & 26.55 \\
LanDiff~\cite{landiff} & 95.41 & \textbf{97.88} & 97.38 & 72.86 & 60.96 & 63.00 & 27.29 \\
\hline
\textbf{\ours (Ours)} & \textbf{97.77} & 97.39 & \textbf{99.10} & 49.89 & \textbf{61.24} & \textbf{66.86} & \textbf{29.01} \\
\bottomrule
\end{tabular*}
\end{table}

\noindent\textbf{Implementation details.}
(1)~Tokenizer: 100K steps on 128 A800 GPUs with a global batch size of 2,048 frame pairs, and $\tau\!=\!0.03$ in \cref{eq:target}, set empirically.
(2)~AR model: Qwen3-VL-8B~\cite{qwen3vl}. Each sample is a chat in which the user turn holds the first frame and the prompt ``Text caption: \{caption\}. Predict the motion token sequence.'', where the caption is empty for unlabeled videos, and the assistant turn holds the delta tokens between two new special tokens, \texttt{<bom>} and \texttt{<eom>}. V2V pretraining runs for 20K steps on 64 A800 GPUs with a global batch size of 768, excluding \texttt{<bom>} and the first 96 tokens from the loss. Trajectory distillation runs for 7K steps on 32 GPUs with a global batch size of 384 and a learning rate of $10^{-6}$, where, in a five-second clip, the student starts from the keyframe at 2\,s with the ground-truth tokens of the next second and samples the tokens of the last 2\,s, and the teacher additionally sees the ground-truth tokens from 0\,s. TI2V fine-tuning runs for 8K steps with the same setting as V2V pretraining.
(3)~Renderer: Wan2.2-TI2V-5B~\cite{wan22} with LoRA of rank 32, trained for 56K steps on 64 A800 GPUs with a global batch size of 64. At inference, we use $s_I\!=\!3$ and $s_T\!=\!5$ in \cref{eq:cfg}.

\noindent\textbf{Evaluation.}
(1)~V2V generation: WorldBench~\cite{worldbench} measures the foreground mIoU of continued videos in four categories, and we evaluate continuations of 33 and 121 frames.
(2)~TI2V generation: Physics-IQ Verified~\cite{physicsiq}, PhyGround~\cite{phyground}, and WorldModelBench~\cite{worldmodelbench} evaluate physical plausibility, and VBench-I2V~\cite{vbenchpp} evaluates general video quality.
(3)~Long video generation: VBench-Long~\cite{vbenchpp} evaluates long videos with its official prompts and lengths. The AR model predicts the delta tokens of the whole video, and the renderer generates it segment by segment, each segment starting from the last frame of the previous one.
(4)~Zero-shot editing: FiVE~\cite{five} contains 420 object-level editing pairs across six editing types. For a fair comparison, we follow AnyV2V~\cite{anyv2v} and edit the first frame with InstructPix2Pix~\cite{instructpix2pix} using the instructions provided by FiVE, and all methods with the first frame as input use this edited frame. We report FiVE-Acc for editing success, the CLIP score of the edited region, the motion fidelity score (MFS), and the time per frame. Methods are grouped into those that keep the latents or features of all source frames and those that keep only the edited first frame and a compact motion representation. For the qualitative motion transfer in Figure~\ref{fig:transfer}, we edit the first frames with Qwen-Image~\cite{qwenimage}.
(5)~Temporal control: TempoControl~\cite{tempocontrol} evaluates whether an object appears, or a strong movement happens, at the second specified by the prompt. An object detector checks that the object is absent before the specified time and present after it, giving the absence (Abs) and presence (Pres) accuracies, which together determine the temporal accuracy (Acc). For movement, Acc checks whether the second with the strongest optical flow matches the prompt.
For VBench-Long and TempoControl, which provide only text prompts, the first frames are generated by Wan2.2-TI2V-5B~\cite{wan22}, the base model of our renderer, so that the gains do not come from better first frames of an external model.

%==========================================================
\subsection{Experimental Results}
\label{sec:exp_results}
\label{sec:benchmarks}
\label{sec:editing}
\label{sec:continuation}
\label{sec:temporal}

\begin{table}[t]
\centering
\caption{Zero-shot video editing on FiVE. The best results among methods with the first frame as input are in bold, and the best results among all methods are underlined.}
\label{tab:five}
\fontsize{6.5}{7.5}\selectfont\renewcommand{\arraystretch}{1.15}
\setlength{\tabcolsep}{4pt}
\begin{tabular*}{\linewidth}{@{\extracolsep{\fill}}l|*{4}{c}@{}}
\toprule
Model & Accuracy (\%)$\uparrow$ & CLIP$_{\text{edited}}\!\uparrow$ & Motion Fidelity$\uparrow$ & Time (s)$\downarrow$ \\
\hline
Source videos & --    & 19.87 & 93.76 & -- \\
\hline
\multicolumn{5}{c}{\emph{All latents as input (time measured on H100)}} \\
\hline
TokenFlow     & 27.43 & 21.15 & 89.00 & 8.04 \\
DMT           & 48.42 & \underline{21.44} & 82.30 & 25.98 \\
VidToMe       & 26.77 & 21.05 & \underline{90.06} & 3.25 \\
AnyV2V~\cite{anyv2v} & 38.02 & 19.72 & 60.36 & 6.11 \\
VideoGrain    & 37.23 & 20.31 & 88.57 & 27.12 \\
Pyramid-Edit  & 43.84 & 20.20 & 80.59 & 1.44 \\
Wan-Edit      & 46.97 & 21.23 & 89.43 & 3.07 \\
\hline
\multicolumn{5}{c}{\emph{First frame as input (time measured on A800)}} \\
\hline
VideoWorld~2~\cite{videoworld2} & 50.98 & 19.17 & 31.36 & 2.30 \\
PhiZero~\cite{phizero}          & 52.72 & 20.38 & 67.30 & \underline{\textbf{0.49}} \\
AdaWorld~\cite{adaworld}        & 47.23 & 18.61 & 66.42 & 1.22 \\
\hline
\textbf{\ours (Ours)}           & \underline{\textbf{53.91}} & \textbf{20.95} & \textbf{80.57} & 1.44 \\
\bottomrule
\end{tabular*}
\end{table}

\begin{table*}[t]
\centering
\caption{Temporal control on TempoControl. The best results among methods without injected temporal information are in bold, and the best results among all methods are underlined.}
\label{tab:temporal}
\fontsize{6.5}{7.5}\selectfont\renewcommand{\arraystretch}{1.15}
\setlength{\tabcolsep}{3pt}
\begin{tabularx}{\linewidth}{@{}l|*{3}{Y}|*{3}{Y}|Y@{}}
\toprule
\multirow{2}{*}{Model} & \multicolumn{3}{c|}{\textbf{One object}} & \multicolumn{3}{c|}{\textbf{Two objects}} & \textbf{Movement} \\
 & Accuracy$\uparrow$ & Absence$\uparrow$ & Presence$\uparrow$ & Accuracy$\uparrow$ & Absence$\uparrow$ & Presence$\uparrow$ & Accuracy$\uparrow$ \\
\hline
\multicolumn{8}{c}{\emph{With injected temporal information}} \\
\hline
CogVideoX-2B$_{\text{w.\ TempoControl}}$~\cite{tempocontrol} & 63.75 & 78.41 & 47.62 & -- & -- & -- & -- \\
Wan2.1-1.3B$_{\text{w.\ TempoControl}}$~\cite{tempocontrol} & \underline{83.56} & 87.38 & \underline{79.75} & 53.17 & 57.32 & \underline{49.02} & 54 \\
\hline
\multicolumn{8}{c}{\emph{Without injected temporal information}} \\
\hline
LTX-2-19B~\cite{ltxvideo}  & 74.23 & 85.23 & 62.12 & 39.31 & 64.88 & 16.08 & 28 \\
CogVideoX-2B~\cite{cogvideox}    & 58.15 & 74.77 & 39.88 & --    & --    & --    & -- \\
CogVideoX1.5-5B~\cite{cogvideox} & 61.43 & 48.64 & \textbf{75.50} & 24.68 & 46.46 & 4.88 & 21 \\
Wan2.1-1.3B~\cite{wan22}     & 63.94 & 67.38 & 60.50 & 37.50 & 45.85 & 29.15 & 19 \\
Wan2.1-14B~\cite{wan22}      & 64.23 & 55.34 & 74.00 & 40.07 & 54.15 & 27.27 & 18 \\
Wan2.2-5B~\cite{wan22}       & 78.02 & 87.42 & 68.00 & 47.05 & 56.86 & \textbf{36.59} & 23 \\
PhiZero~\cite{phizero} & 50.00 & \underline{\textbf{100.0}} & 0.00 & 37.51 & 77.80 & 0.89 & 24 \\
\hline
\textbf{\ours (Ours)} & \textbf{79.19} & 99.38 & 59.00 & \underline{\textbf{54.65}} & \underline{\textbf{83.05}} & 28.82 & \underline{\textbf{61}} \\
\bottomrule
\end{tabularx}
\end{table*}

\noindent\textbf{V2V generation.}
As shown in Table~\ref{tab:worldbench}, \ours already outperforms all Cosmos models after pretraining, and trajectory distillation further raises the overall score by 0.075 at 33 frames and by 0.114 at 121 frames, reaching 0.620 and 0.480.
The larger gain at 121 frames agrees with errors compounding over long rollouts~\cite{worldbench}, where training on its own rollouts matters more.
The student rolls out from a short history and may make errors, while the teacher sees a longer ground-truth history. By fitting the distributions of the teacher on its own rollouts, the student learns to correct its errors, which reduces this accumulation, an advantage enabled by predicting discrete delta tokens.

\noindent\textbf{TI2V generation.}
We evaluate TI2V generation in two aspects, physical plausibility and general video quality.
\begin{itemize}[leftmargin=*,nosep]
\item \textbf{Physical plausibility.} As shown in Table~\ref{tab:physics}, \ours achieves the best overall scores on all three physics benchmarks. On Physics-IQ, \ours reaches an IQ-Score of 48.1, 6.9 points above PhiZero, and its ST-IoU of 65.0 nearly doubles that of PhiZero (36.8), showing that the motion happens not only at the right place but also at the right time. Because each chunk of delta tokens describes one step, the AR model predicts when each event happens, which clip-level tokens model poorly. PhiZero has a higher S-IoU, but this metric ignores when the motion happens. On PhyGround and WorldModelBench, \ours achieves overall scores of 3.10 and 8.66, with the largest gains in physics (3.12 vs.\ 3.01) and common sense (1.91 vs.\ 1.71) over PhiZero.
\item \textbf{General video quality.} As shown in Table~\ref{tab:vbench}, \ours achieves the best I2V score of 97.80 on VBench-I2V, and its total score of 89.12 is second only to Magi-1 with 24B parameters. \Cref{tab:vbench_i2v} in the appendix reports all dimensions of VBench-I2V.
\end{itemize}

\noindent\textbf{Long video generation.}
As shown in Table~\ref{tab:vbench_long}, \ours achieves the best subject consistency, motion smoothness, aesthetic and imaging quality, and overall consistency on VBench-Long, outperforming ARLON and LanDiff, which predict the state of every frame or keyframe, and Figure~\ref{fig:long} shows that it keeps the subject and the scene consistent over the whole video.
Since each chunk is encoded independently, generation continues chunk by chunk without drifting in appearance.
LanDiff reaches a higher dynamic degree, as its videos show the color and content drift in Figure~\ref{fig:long}, which optical flow also counts as motion.
StreamingT2V reaches the highest dynamic degree but the lowest subject consistency and motion smoothness.
The dynamic degree of \ours (49.89) is comparable to those of ARLON and OpenSora-V1.2 and higher than that of FreeNoise.

\noindent\textbf{Zero-shot editing.}
As shown in Table~\ref{tab:five} and Figure~\ref{fig:five}, \ours achieves the highest FiVE-Acc of 53.91 and the highest MFS of 80.57 among methods with the first frame as input.
Its FiVE-Acc is also slightly above those of DMT and Wan-Edit, which keep all latents of the source video.
Methods with all latents preserve more motion as they keep every source frame. Among methods that keep only the edited first frame and a compact motion representation, the delta tokens from \ours carry more of the source motion than those of PhiZero and AdaWorld.
Methods with the first frame as input need no inversion of the source video and are generally faster. \ours takes 1.44\,s per frame on an A800, slightly slower than PhiZero and AdaWorld by 0.95\,s and 0.22\,s, and on par with the fastest method with all latents measured on an H100.
We further test motion transfer on more open-domain videos. As shown in Figure~\ref{fig:transfer}, \ours faithfully transfers the motion of the source videos to new subjects.

%==========================================================

\noindent\textbf{Temporal control.}
As shown in Table~\ref{tab:temporal}, \ours achieves the best accuracy for two objects (54.65) and movement (61) and is second only to the method with injected temporal information for one object (79.19).
PhiZero reaches a presence accuracy of only 0.00 and 0.89, which suggests that its tokens barely model temporal information in these scenarios.
Figure~\ref{fig:tempo_vis} shows that the backpack, the cat, and the strong movements of the child and the wolf appear at the times specified by the prompts, as each chunk of delta tokens covers one third of a second and lets the AR model place events at this granularity.

\begin{table*}[t]
\centering
\caption{Ablation of the tokenizer training objective. Token flip rate (\%)$\downarrow$ on Inter4K under four levels of each perturbation, from the weakest (Level 1) to the strongest (Level 4).}
\label{tab:abl_tok}
\fontsize{6.5}{7.5}\selectfont\renewcommand{\arraystretch}{1.15}
\setlength{\tabcolsep}{2pt}
\begin{tabularx}{\linewidth}{@{}l|*{4}{Y}|*{4}{Y}|*{4}{Y}|*{4}{Y}|Y@{}}
\toprule
\multirow{2}{*}{Model} & \multicolumn{4}{c|}{\textbf{H.264 (CRF)}} & \multicolumn{4}{c|}{\textbf{Shot noise}} & \multicolumn{4}{c|}{\textbf{Brightness}} & \multicolumn{4}{c|}{\textbf{Contrast}} & \multirow{2}{*}{\textbf{Average}} \\
 & Level 1 & Level 2 & Level 3 & Level 4 & Level 1 & Level 2 & Level 3 & Level 4 & Level 1 & Level 2 & Level 3 & Level 4 & Level 1 & Level 2 & Level 3 & Level 4 & \\
\hline
X-Omni & 33.3 & 43.1 & 52.3 & 65.3 & 31.1 & 37.2 & 46.1 & 55.4 & 15.1 & 18.7 & 24.0 & 28.0 & 16.0 & 20.0 & 23.3 & 26.0 & 33.4 \\
\hline
\ours$_{\text{w.\ MSE}}$ & 77.3 & 83.5 & 88.9 & 94.9 & 75.5 & 78.9 & 83.7 & 88.2 & 66.9 & 70.4 & 74.6 & 77.5 & 69.5 & 71.9 & 74.0 & 75.9 & 78.2 \\
\textbf{\ours$_{\text{w.\ KL}}$ (Ours)} & \textbf{47.0} & \textbf{61.7} & \textbf{74.3} & \textbf{87.5} & \textbf{41.1} & \textbf{50.3} & \textbf{63.0} & \textbf{74.9} & \textbf{27.2} & \textbf{34.4} & \textbf{44.9} & \textbf{52.2} & \textbf{23.8} & \textbf{32.3} & \textbf{39.2} & \textbf{44.8} & \textbf{49.9} \\
\bottomrule
\end{tabularx}
\end{table*}

\begin{table}[t]
\centering
\caption{Ablation of V2V training on TI2V generation.}
\label{tab:abl_v2v}
\fontsize{6.5}{7.5}\selectfont\renewcommand{\arraystretch}{1.15}
\setlength{\tabcolsep}{2pt}
\begin{tabular*}{\linewidth}{@{\extracolsep{\fill}}l|*{4}{c}@{}}
\toprule
Model & Physics-IQ$\uparrow$ & PhyGround$\uparrow$ & WorldModelBench$\uparrow$ & VBench-I2V$\uparrow$ \\
\hline
\ours$_{\text{w/o.\ V2V}}$ & \textbf{48.1} & 3.02 & 8.37 & 88.94 \\
\textbf{\ours$_{\text{w.\ V2V}}$ (Ours)} & \textbf{48.1} & \textbf{3.10} & \textbf{8.66} & \textbf{89.12} \\
\bottomrule
\end{tabular*}
\end{table}

\begin{table}[t]
\centering
\caption{Ablation of the source of motion in rendering on FiVE. Full: all conditions. w/o.~$c_I$: drop visual condition. w.~SigLIP2: trained with only the SigLIP2-g features of the first frame.}
\label{tab:abl_edit}
\fontsize{6.5}{7.5}\selectfont\renewcommand{\arraystretch}{1.15}
\setlength{\tabcolsep}{4pt}
\begin{tabular*}{\linewidth}{@{\extracolsep{\fill}}l|*{3}{c}|*{3}{c}@{}}
\toprule
\multirow{2}{*}{Model} & \multicolumn{3}{c|}{Motion Fidelity$_{\text{whole}}\!\uparrow$} & \multicolumn{3}{c}{Motion Fidelity$_{\text{edited}}\!\uparrow$} \\
 & Full & w/o.~$c_I$ & $\Delta$ & Full & w/o.~$c_I$ & $\Delta$ \\
\hline
VideoWorld~2~\cite{videoworld2} & 31.36 & 30.17 & $-$1.19 & 31.04 & 28.75 & $-$2.30 \\
PhiZero~\cite{phizero} & 67.30 & 49.90 & $-$17.40 & 64.61 & 46.50 & $-$18.11 \\
AdaWorld~\cite{adaworld} & 66.42 & 35.36 & $-$31.06 & 62.85 & 32.52 & $-$30.33 \\
Wan2.2-5B$_{\text{w.\ SigLIP2}}$ & 54.69 & -- & -- & 51.07 & -- & -- \\
\hline
\textbf{\ours (Ours)} & \textbf{80.57} & 55.78 & $-$24.78 & \textbf{77.99} & 51.26 & $-$26.72 \\
\bottomrule
\end{tabular*}
\end{table}

\subsection{Ablation Study}
\label{sec:ablation}
\label{sec:exp_tok}

\noindent\textbf{Tokenizer training objective.}
Under the same tokenizer architecture, we compare two training objectives: predicting distributions and regressing features with an MSE loss.
We test whether the tokens follow semantics by adding slight perturbations to the frame pairs of the source videos and measuring the token flip rate, \ie, the fraction of tokens that change.
We choose the 1,000 4K videos of Inter4K~\cite{inter4k}, so that the source videos themselves carry little such noise, and apply the four perturbations in Table~\ref{tab:abl_tok} independently to each frame at four levels each, following~\cite{yi2021robustness}.
As shown in Table~\ref{tab:abl_tok}, predicting distributions lowers the flip rate on average and under every perturbation, narrowing the gap to the semantic tokens of X-Omni.
Even at Level 1, where the PSNR stays above 41\,dB, regression flips the majority of tokens.
However, by predicting a distribution over semantic tokens, our tokenizer largely ignores such subtle changes.

\noindent\textbf{V2V training for TI2V.}
Here, V2V training refers to V2V pretraining and trajectory distillation.
For comparison, we train the AR model directly for TI2V from Qwen3-VL-8B on the same 2.0M captioned clips for the same number of training steps.
As shown in Table~\ref{tab:abl_v2v}, V2V training improves PhyGround from 3.02 to 3.10, WorldModelBench from 8.37 to 8.66, and VBench-I2V from 88.94 to 89.12.
Pretraining on abundant unlabeled videos provides a better starting point for the downstream TI2V task.

\noindent\textbf{Source of motion in rendering.}
We examine where the motion that the renderer preserves comes from: the text, the SigLIP2-g features, or the delta tokens.
We evaluate it using the motion fidelity score on FiVE.
In Table~\ref{tab:abl_edit}, we first drop the visual condition $c_I$ at inference, \ie, the SigLIP2-g features together with the motion representation, as our renderer also does during training.
Every method falls to an MFS between 30 and 56.
Delta tokens contribute 24.78 points, more than the tokens of PhiZero and the codes of VideoWorld~2, and only the latent actions of AdaWorld contribute more, but from a lower starting point.
Dropping $c_I$ cannot separate the SigLIP2-g features and the delta tokens, so we train another renderer with only the SigLIP2-g features of the first frame for the same 56K steps, identical to ours except for the delta tokens.
It reaches an MFS of only 54.69 and 51.07, close to those of \ours without $c_I$ (55.78 and 51.26), so the motion that the renderer preserves comes from the delta tokens.

\FloatBarrier
\section{Conclusion}
\label{sec:conclusion}

We presented \ours, which decouples the AR reasoner from the diffusion renderer.
The AR model predicts semantic delta tokens, which encode how frames evolve in the SigLIP2 feature space, and a diffusion renderer generates the video from them chunk by chunk.
\ours shows immense potential in V2V generation and physics-aware TI2V generation, generates long videos well beyond its five-second training clips, transfers motion zero-shot, and follows timing instructions at a given second, with visual quality comparable to strong video generation models.

{
    \small
    \bibliographystyle{ieeenat_fullname}
    \bibliography{main,appendix}

@inproceedings{hoigen,
  title={Hoigen-1m: A large-scale dataset for human-object interaction video generation},
  author={Liu, Kun and Liu, Qi and Liu, Xinchen and Li, Jie and Zhang, Yongdong and Luo, Jiebo and He, Xiaodong and Liu, Wu},
  booktitle={2025 IEEE/CVF Conference on Computer Vision and Pattern Recognition (CVPR)},
  pages={24001--24010},
  year={2025},
  organization={IEEE}
}

@inproceedings{openvid,
  title={Openvid-1m: A large-scale high-quality dataset for text-to-video generation},
  author={Nan, Kepan and Xie, Rui and Zhou, Penghao and Fan, Tiehan and Yang, Zhenheng and Chen, Zhijie and Li, Xiang and Yang, Jian and Tai, Ying},
  booktitle={International conference on learning representations},
  volume={2025},
  pages={1045--1064},
  year={2025}
}

@article{sekai,
  title={Sekai: A video dataset towards world exploration},
  author={Li, Zhen and Li, Chuanhao and Mao, Xiaofeng and Lin, Shaoheng and Li, Ming and Zhao, Shitian and Xu, Zhaopan and Li, Xinyue and Feng, Yukang and Sun, Jianwen and others},
  journal={Advances in Neural Information Processing Systems},
  volume={38},
  year={2026}
}

@article{ultravideo,
  title={Ultravideo: High-quality uhd video dataset with comprehensive captions},
  author={Xue, Zhucun and Zhang, Jiangning and Hu, Teng and He, Haoyang and Chen, Yinan and Cai, Yuxuan and Wang, Yabiao and Wang, Chengjie and Liu, Yong and Li, Xiangtai and others},
  journal={arXiv preprint arXiv:2506.13691},
  year={2025}
}

@article{wisa,
  title={Wisa: World simulator assistant for physics-aware text-to-video generation},
  author={Wang, Jing and Ma, Ao and Cao, Ke and Zheng, Jun and Feng, Jiasong and Zhang, Zhanjie and Pang, Wanyuan and Liang, Xiaodan},
  journal={Advances in Neural Information Processing Systems},
  volume={38},
  pages={5388--5416},
  year={2026}
}

@article{momentsintime,
  title={Moments in time dataset: one million videos for event understanding},
  author={Monfort, Mathew and Andonian, Alex and Zhou, Bolei and Ramakrishnan, Kandan and Bargal, Sarah Adel and Yan, Tom and Brown, Lisa and Fan, Quanfu and Gutfreund, Dan and Vondrick, Carl and others},
  journal={IEEE transactions on pattern analysis and machine intelligence},
  volume={42},
  number={2},
  pages={502--508},
  year={2019},
  publisher={IEEE}
}

@article{comphy,
  title={Comphy: Compositional physical reasoning of objects and events from videos},
  author={Chen, Zhenfang and Yi, Kexin and Li, Yunzhu and Ding, Mingyu and Torralba, Antonio and Tenenbaum, Joshua B and Gan, Chuang},
  journal={arXiv preprint arXiv:2205.01089},
  year={2022}
}

@article{physion,
  title={Physion: Evaluating physical prediction from vision in humans and machines},
  author={Bear, Daniel M and Wang, Elias and Mrowca, Damian and Binder, Felix J and Tung, Hsiao-Yu Fish and Pramod, RT and Holdaway, Cameron and Tao, Sirui and Smith, Kevin and Sun, Fan-Yun and others},
  journal={arXiv preprint arXiv:2106.08261},
  year={2021}
}

@article{physionpp,
  title={Physion++: Evaluating physical scene understanding that requires online inference of different physical properties},
  author={Tung, Hsiao-Yu and Ding, Mingyu and Chen, Zhenfang and Bear, Daniel and Gan, Chuang and Tenenbaum, Josh and Yamins, Dan and Fan, Judith and Smith, Kevin},
  journal={Advances in Neural Information Processing Systems},
  volume={36},
  pages={67048--67068},
  year={2023}
}

@article{clevrer,
  title={Clevrer: Collision events for video representation and reasoning},
  author={Yi, Kexin and Gan, Chuang and Li, Yunzhu and Kohli, Pushmeet and Wu, Jiajun and Torralba, Antonio and Tenenbaum, Joshua B},
  journal={arXiv preprint arXiv:1910.01442},
  year={2019}
}

@article{phyco,
  title={PhyCo: Learning Controllable Physical Priors for Generative Motion},
  author={Narayanan, Sriram and Jiang, Ziyu and Narasimhan, Srinivasa and Chandraker, Manmohan},
  journal={arXiv preprint arXiv:2604.28169},
  year={2026}
}

@article{vidgen,
  title={Vidgen-1m: A large-scale dataset for text-to-video generation},
  author={Tan, Zhiyu and Yang, Xiaomeng and Qin, Luozheng and Li, Hao},
  journal={arXiv preprint arXiv:2408.02629},
  year={2024}
}

@inproceedings{egodex,
  title={Egodex: Learning dexterous manipulation from large-scale egocentric video},
  author={Hoque, Ryan and Huang, Peide and Yoon, David and Zhang, Jian and others},
  booktitle={International Conference on Learning Representations},
  volume={2026},
  pages={4218--4237},
  year={2026}
}

@article{llavaov2,
  title={Llava-onevision-2: Towards next-generation perceptual intelligence},
  author={An, Xiang and Xie, Yin and Tang, Feilong and Yan, Yunyao and Tan, Huajie and Zhu, Didi and Chen, Changrui and Zhao, Xiuwei and Qin, Bin and Yang, Kaicheng and others},
  journal={arXiv preprint arXiv:2605.25979},
  year={2026}
}

@article{kinetics700,
  title={A short note on the kinetics-700 human action dataset},
  author={Carreira, Joao and Noland, Eric and Hillier, Chloe and Zisserman, Andrew},
  journal={arXiv preprint arXiv:1907.06987},
  year={2019}
}

@article{kinetics400,
  title={The kinetics human action video dataset},
  author={Kay, Will and Carreira, Joao and Simonyan, Karen and Zhang, Brian and Hillier, Chloe and Vijayanarasimhan, Sudheendra and Viola, Fabio and Green, Tim and Back, Trevor and Natsev, Paul and others},
  journal={arXiv preprint arXiv:1705.06950},
  year={2017}
}

@inproceedings{epickitchens,
  title={Scaling egocentric vision: the dataset},
  author={Damen, Dima and Doughty, Hazel and Farinella, Giovanni Maria and Fidler, Sanja and Furnari, Antonino and Kazakos, Evangelos and Moltisanti, Davide and Munro, Jonathan and Perrett, Toby and Price, Will and others},
  booktitle={European conference on computer vision},
  pages={753--771},
  year={2018},
  organization={Springer}
}

@inproceedings{oops,
  title={Oops! predicting unintentional action in video},
  author={Epstein, Dave and Chen, Boyuan and Vondrick, Carl},
  booktitle={2020 IEEE/CVF Conference on Computer Vision and Pattern Recognition (CVPR)},
  pages={916--926},
  year={2020},
  organization={IEEE}
}

@inproceedings{motionbench,
  title={Motionbench: Benchmarking and improving fine-grained video motion understanding for vision language models},
  author={Hong, Wenyi and Cheng, Yean and Yang, Zhuoyi and Wang, Weihan and Wang, Lefan and Gu, Xiaotao and Huang, Shiyu and Dong, Yuxiao and Tang, Jie},
  booktitle={2025 IEEE/CVF Conference on Computer Vision and Pattern Recognition (CVPR)},
  pages={8450--8460},
  year={2025},
  organization={IEEE}
}

@inproceedings{multisports,
  title={Multisports: A multi-person video dataset of spatio-temporally localized sports actions},
  author={Li, Yixuan and Chen, Lei and He, Runyu and Wang, Zhenzhi and Wu, Gangshan and Wang, Limin},
  booktitle={2021 IEEE/CVF International Conference on Computer Vision (ICCV)},
  pages={13516--13525},
  year={2021},
  organization={IEEE}
}

@article{temporalbench,
  title={Temporalbench: Benchmarking fine-grained temporal understanding for multimodal video models},
  author={Cai, Mu and Tan, Reuben and Zhang, Jianrui and Zou, Bocheng and Zhang, Kai and Yao, Feng and Zhu, Fangrui and Gu, Jing and Zhong, Yiwu and Shang, Yuzhang and others},
  journal={arXiv preprint arXiv:2410.10818},
  year={2024}
}

@inproceedings{kubric,
  title={Kubric: A scalable dataset generator},
  author={Greff, Klaus and Belletti, Francois and Beyer, Lucas and Doersch, Carl and Du, Yilun and Duckworth, Daniel and Fleet, David J and Gnanapragasam, Dan and Golemo, Florian and Herrmann, Charles and others},
  booktitle={Proceedings of the IEEE/CVF conference on computer vision and pattern recognition},
  pages={3749--3761},
  year={2022}
}

@inproceedings{galaxea,
  title={Galaxy open-world dataset and g0 dual-system vla model},
  author={Jiang, Tao and Yuan, Tianyuan and Liu, Yicheng and Lu, Chenhao and Cui, Jianning and Liu, Xiao and Cheng, Shuiqi and Gao, Jiyang and Xu, Huazhe and Zhao, Hang},
  booktitle={2026 IEEE International Conference on Robotics and Automation (ICRA)},
  pages={20749--20756},
  year={2026},
  organization={IEEE}
}

@inproceedings{activitynet,
  title={Activitynet: A large-scale video benchmark for human activity understanding},
  author={Caba Heilbron, Fabian and Escorcia, Victor and Ghanem, Bernard and Carlos Niebles, Juan},
  booktitle={Proceedings of the ieee conference on computer vision and pattern recognition},
  pages={961--970},
  year={2015}
}

@article{physar,
  title={Reasoning physical video generation with diffusion timestep tokens via reinforcement learning},
  author={Lin, Wang and Jia, Liyu and Hu, Wentao and Pan, Kaihang and Yue, Zhongqi and Zhao, Wei and Chen, Jingyuan and Wu, Fei and Zhang, Hanwang},
  journal={arXiv preprint arXiv:2504.15932},
  year={2025}
}

@article{stepvideoti2v,
  title={Step-video-ti2v technical report: A state-of-the-art text-driven image-to-video generation model},
  author={Huang, Haoyang and Ma, Guoqing and Duan, Nan and Chen, Xing and Wan, Changyi and Ming, Ranchen and Wang, Tianyu and Wang, Bo and Lu, Zhiying and Li, Aojie and others},
  journal={arXiv preprint arXiv:2503.11251},
  year={2025}
}

@inproceedings{seine,
  title={Seine: Short-to-long video diffusion model for generative transition and prediction},
  author={Chen, Xinyuan and Wang, Yaohui and Zhang, Lingjun and Zhuang, Shaobin and Ma, Xin and Yu, Jiashuo and Wang, Yali and Lin, Dahua and Qiao, Yu and Liu, Ziwei},
  booktitle={The Twelfth International Conference on Learning Representations},
  year={2023}
}

@article{i2vgenxl,
  title={I2vgen-xl: High-quality image-to-video synthesis via cascaded diffusion models},
  author={Zhang, Shiwei and Wang, Jiayu and Zhang, Yingya and Zhao, Kang and Yuan, Hangjie and Qin, Zhiwu and Wang, Xiang and Zhao, Deli and Zhou, Jingren},
  journal={arXiv preprint arXiv:2311.04145},
  year={2023}
}

@article{consisti2v,
  title={Consisti2v: Enhancing visual consistency for image-to-video generation},
  author={Ren, Weiming and Yang, Huan and Zhang, Ge and Wei, Cong and Du, Xinrun and Huang, Wenhao and Chen, Wenhu},
  journal={arXiv preprint arXiv:2402.04324},
  year={2024}
}

@inproceedings{videocrafter2,
  title={Videocrafter2: Overcoming data limitations for high-quality video diffusion models},
  author={Chen, Haoxin and Zhang, Yong and Cun, Xiaodong and Xia, Menghan and Wang, Xintao and Weng, Chao and Shan, Ying},
  booktitle={2024 IEEE/CVF Conference on Computer Vision and Pattern Recognition (CVPR)},
  pages={7310--7320},
  year={2024},
  organization={IEEE}
}

@article{vit,
  title={An image is worth 16x16 words: Transformers for image recognition at scale},
  author={Dosovitskiy, Alexey},
  journal={arXiv preprint arXiv:2010.11929},
  year={2020}
}

@article{laionbvd,
  title={LAION-BVD: A 10-Million-Hour Open Video Dataset for Multimodal Pre-training},
  author={Hochlehnert, Andreas and Nezhurina, Marianna and Cherti, Mehdi and Radonjic, Andrej and Wiedemer, Thadd{\"a}us and Schuhmann, Christoph and Beaumont, Romain and Brendel, Wieland and Sch{\"o}lkopf, Bernhard and Koepke, A and others},
  journal={arXiv preprint arXiv:2608.24845},
  year={2026}
}

@String(CVPR  = {CVPR})

@String(ICCV  = {ICCV})

@String(AAAI  = {AAAI})

@article{phizero,
  title={Phizero: A world model built around physical language},
  author={Shang, Shuyao and Wang, Yuqi and Gao, Ruopeng and Chen, Xu and Tan, Tieniu and Fan, Lue and Zhang, Zhaoxiang},
  journal={arXiv preprint arXiv:2607.28624},
  year={2026}
}

@article{siglip2,
  title={Siglip 2: Multilingual vision-language encoders with improved semantic understanding, localization, and dense features},
  author={Tschannen, Michael and Gritsenko, Alexey and Wang, Xiao and Naeem, Muhammad Ferjad and Alabdulmohsin, Ibrahim and Parthasarathy, Nikhil and Evans, Talfan and Beyer, Lucas and Xia, Ye and Mustafa, Basil and others},
  journal={arXiv preprint arXiv:2502.14786},
  year={2025}
}

@article{xomni,
  title={X-omni: Reinforcement learning makes discrete autoregressive image generative models great again},
  author={Geng, Zigang and Wang, Yibing and Ma, Yeyao and Li, Chen and Rao, Yongming and Gu, Shuyang and Zhong, Zhao and Lu, Qinglin and Hu, Han and Zhang, Xiaosong and others},
  journal={arXiv preprint arXiv:2507.22058},
  year={2025}
}

@inproceedings{fsq,
  title={Finite scalar quantization: Vq-vae made simple},
  author={Mentzer, Fabian and Minnen, David and Agustsson, Eirikur and Tschannen, Michael},
  booktitle={International Conference on Learning Representations},
  volume={2024},
  pages={51772--51783},
  year={2024}
}

@article{wan22,
  title={Wan: Open and advanced large-scale video generative models},
  author={Wan, Team and Wang, Ang and Ai, Baole and Wen, Bin and Mao, Chaojie and Xie, Chen-Wei and Chen, Di and Yu, Feiwu and Zhao, Haiming and Yang, Jianxiao and others},
  journal={arXiv preprint arXiv:2503.20314},
  year={2025}
}

@article{qwen3vl,
  title={Qwen3-vl technical report},
  author={Bai, Shuai and Cai, Yuxuan and Chen, Ruizhe and Chen, Keqin and Chen, Xionghui and Cheng, Zesen and Deng, Lianghao and Ding, Wei and Gao, Chang and Ge, Chunjiang and others},
  journal={arXiv preprint arXiv:2511.21631},
  year={2025}
}

@inproceedings{blip2,
  title={Blip-2: Bootstrapping language-image pre-training with frozen image encoders and large language models},
  author={Li, Junnan and Li, Dongxu and Savarese, Silvio and Hoi, Steven},
  booktitle={International conference on machine learning},
  pages={19730--19742},
  year={2023},
  organization={PmLR}
}

@article{physicsiq,
  title={Physics-IQ Verified},
  author={R{\"a}dsch, Tim and Asano, Yuki M and Kuehne, Hilde and Bauer, Stefan and Jaini, Priyank and Geirhos, Robert and L{\"u}th, Carsten T},
  journal={arXiv preprint arXiv:2606.18943},
  year={2026}
}

@article{phyground,
  title={PhyGround: Benchmarking Physical Reasoning in Generative World Models},
  author={Lin, Juyi and Akbari, Arash and He, Yumei and Zhao, Lin and Zhang, Haichao and Akbari, Arman and Xu, Xingchen and Lu, Zoe Y and Nan, Enfu and Deng, Hokin and others},
  journal={arXiv preprint arXiv:2605.10806},
  year={2026}
}

@article{worldmodelbench,
  title={Worldmodelbench: Judging video generation models as world models},
  author={Li, Dacheng and Fang, Yunhao and Chen, Yukang and Yang, Shuo and Cao, Shiyi and Wong, Justin and Luo, Michael and Wang, Xiaolong and Yin, Hongxu and Gonzalez, Joseph and others},
  journal={Advances in Neural Information Processing Systems},
  volume={38},
  year={2026}
}

@article{vbenchpp,
  title={Vbench++: Comprehensive and versatile benchmark suite for video generative models},
  author={Huang, Ziqi and Zhang, Fan and Xu, Xiaojie and He, Yinan and Yu, Jiashuo and Dong, Ziyue and Ma, Qianli and Chanpaisit, Nattapol and Si, Chenyang and Jiang, Yuming and others},
  journal={IEEE Transactions on Pattern Analysis and Machine Intelligence},
  year={2025},
  publisher={IEEE}
}

@inproceedings{five,
  title={Five-bench: A fine-grained video editing benchmark for evaluating emerging diffusion and rectified flow models},
  author={Li, Minghan and Xie, Chenxi and Wu, Yichen and Zhang, Lei and Wang, Mengyu},
  booktitle={2025 IEEE/CVF International Conference on Computer Vision (ICCV)},
  pages={16672--16681},
  year={2025},
  organization={IEEE}
}

@inproceedings{streamingt2v,
  title={Streamingt2v: Consistent, dynamic, and extendable long video generation from text},
  author={Henschel, Roberto and Khachatryan, Levon and Poghosyan, Hayk and Hayrapetyan, Daniil and Tadevosyan, Vahram and Wang, Zhangyang and Navasardyan, Shant and Shi, Humphrey},
  booktitle={2025 IEEE/CVF Conference on Computer Vision and Pattern Recognition (CVPR)},
  pages={2568--2577},
  year={2025},
  organization={IEEE}
}

@article{fifodiffusion,
  title={Fifo-diffusion: Generating infinite videos from text without training},
  author={Kim, Jihwan and Kang, Junoh and Choi, Jinyoung and Han, Bohyung},
  journal={Advances in Neural Information Processing Systems},
  volume={37},
  pages={89834--89868},
  year={2024}
}

@article{framepack,
  title={Frame context packing and drift prevention in next-frame-prediction video diffusion models},
  author={Zhang, Lvmin and Cai, Shengqu and Li, Muyang and Wetzstein, Gordon and Agrawala, Maneesh},
  journal={Advances in Neural Information Processing Systems},
  volume={38},
  pages={30546--30566},
  year={2026}
}

@article{skyreelsv2,
  title={Skyreels-v2: Infinite-length film generative model},
  author={Chen, Guibin and Lin, Dixuan and Yang, Jiangping and Lin, Chunze and Zhu, Junchen and Fan, Mingyuan and Zhang, Hao and Chen, Sheng and Chen, Zheng and Ma, Chengcheng and others},
  journal={arXiv preprint arXiv:2504.13074},
  year={2025}
}

@inproceedings{selfforcingpp,
  title={Self-forcing++: Towards minute-scale high-quality video generation},
  author={Cui, Jiaxing and Wu, Jie and Li, Ming and Yang, Tao and Li, Xiaojie and Wang, Rui and Bai, Andrew and Ban, Yuanhao and Hsieh, Cho-Jui},
  booktitle={International Conference on Learning Representations},
  volume={2026},
  pages={85802--85822},
  year={2026}
}

@article{longlive,
  title={Longlive: Real-time interactive long video generation},
  author={Yang, Shuai and Huang, Wei and Chu, Ruihang and Xiao, Yicheng and Zhao, Yuyang and Wang, Xianbang and Li, Muyang and Xie, Enze and Chen, Yingcong and Lu, Yao and others},
  journal={arXiv preprint arXiv:2509.22622},
  year={2025}
}

@article{diffusionforcing,
  title={Diffusion forcing: Next-token prediction meets full-sequence diffusion},
  author={Chen, Boyuan and Mart{\'\i} Mons{\'o}, Diego and Du, Yilun and Simchowitz, Max and Tedrake, Russ and Sitzmann, Vincent},
  journal={Advances in Neural Information Processing Systems},
  volume={37},
  pages={24081--24125},
  year={2024}
}

@article{historyguidance,
  title={History-guided video diffusion},
  author={Song, Kiwhan and Chen, Boyuan and Simchowitz, Max and Du, Yilun and Tedrake, Russ and Sitzmann, Vincent},
  journal={arXiv preprint arXiv:2502.06764},
  year={2025}
}

@inproceedings{causvid,
  title={From slow bidirectional to fast autoregressive video diffusion models},
  author={Yin, Tianwei and Zhang, Qiang and Zhang, Richard and Freeman, William T and Durand, Fredo and Shechtman, Eli and Huang, Xun},
  booktitle={2025 IEEE/CVF Conference on Computer Vision and Pattern Recognition (CVPR)},
  pages={22963--22974},
  year={2025},
  organization={IEEE}
}

@article{selfforcing,
  title={Self forcing: Bridging the train-test gap in autoregressive video diffusion},
  author={Huang, Xun and Li, Zhengqi and He, Guande and Zhou, Mingyuan and Shechtman, Eli},
  journal={Advances in Neural Information Processing Systems},
  volume={38},
  pages={167283--167308},
  year={2026}
}

@article{magi1,
  title={Magi-1: Autoregressive video generation at scale},
  author={Teng, Hansi and Jia, Hongyu and Sun, Lei and Li, Lingzhi and Li, Maolin and Tang, Mingqiu and Han, Shuai and Zhang, Tianning and Zhang, WQ and Luo, Weifeng and others},
  journal={arXiv preprint arXiv:2505.13211},
  year={2025}
}

@inproceedings{nova,
  title={Autoregressive video generation without vector quantization},
  author={Deng, Haoge and Pan, Ting and Diao, Haiwen and Luo, Zhengxiong and Cui, Yufeng and Lu, Huchuan and Shan, Shiguang and Qi, Yonggang and Wang, Xinlong},
  booktitle={International Conference on Learning Representations},
  volume={2025},
  pages={44730--44745},
  year={2025}
}

@inproceedings{pyramidflow,
  title={Pyramidal flow matching for efficient video generative modeling},
  author={Jin, Yang and Sun, Zhicheng and Li, Ningyuan and Xu, Kun and Jiang, Hao and Zhuang, Nan and Huang, Quzhe and Song, Yang and Mu, Yadong and Lin, Zhouchen},
  booktitle={International Conference on Learning Representations},
  volume={2025},
  pages={23378--23402},
  year={2025}
}

@article{cosmos,
  title={Cosmos world foundation model platform for physical ai},
  author={Agarwal, Niket and Ali, Arslan and Bala, Maciej and Balaji, Yogesh and Barker, Erik and Cai, Tiffany and Chattopadhyay, Prithvijit and Chen, Yongxin and Cui, Yin and Ding, Yifan and others},
  journal={arXiv preprint arXiv:2501.03575},
  year={2025}
}

@article{cosmos3,
  title={Cosmos 3: Omnimodal world models for physical ai},
  author={Agarwal, Niket and Ali, Arslan and Allen, Jon and Antolini, Martin and Aubame, Adeline and Azzolini, Alisson and Bai, Junjie and Bala, Maciej and Balaji, Yogesh and Bapst, Josh and others},
  journal={arXiv preprint arXiv:2606.02800},
  year={2026}
}

@misc{sora2,
  title        = {Sora 2 System Card},
  author       = {{OpenAI}},
  year         = {2025},
  howpublished = {\url{https://cdn.openai.com/pdf/50d5973c-c4ff-4c2d-986f-c72b5d0ff069/sora_2_system_card.pdf}},
  note         = {Accessed: January 21, 2026}
}

@misc{grokvideo,
  title        = {{Grok Imagine API}: State-of-the-Art Video Generation across Quality, Cost, and Latency},
  author       = {{xAI}},
  year         = {2026},
  howpublished = {\url{https://x.ai/news/grok-imagine-api}},
  note         = {Accessed: April 29, 2026}
}

@article{ltxvideo,
  title={Ltx-2: Efficient joint audio-visual foundation model},
  author={HaCohen, Yoav and Brazowski, Benny and Chiprut, Nisan and Bitterman, Yaki and Kvochko, Andrew and Berkowitz, Avishai and Shalem, Daniel and Lifschitz, Daphna and Moshe, Dudu and Porat, Eitan and others},
  journal={arXiv preprint arXiv:2601.03233},
  year={2026}
}

@inproceedings{cosmos25,
  title={Cosmos world foundation models for physical ai},
  author={Gu, Jinwei},
  booktitle={Proceedings of the 3rd International Workshop on Rich Media With Generative AI},
  pages={39--39},
  year={2025}
}

@article{omniweaving,
  title={Omniweaving: Towards unified video generation with free-form composition and reasoning},
  author={Pan, Kaihang and Tian, Qi and Zhang, Jianwei and Kong, Weijie and Xiong, Jiangfeng and Long, Yanxin and Zhang, Shixue and Qiu, Haiyi and Wang, Tan and Lv, Zheqi and others},
  journal={arXiv preprint arXiv:2603.24458},
  year={2026}
}

@misc{veo31,
  title        = {{Veo 3.1}},
  author       = {{Google DeepMind}},
  year         = {2025},
  howpublished = {\url{https://deepmind.google/models/veo/}},
  note         = {Accessed: April 29, 2026}
}

@article{pandora,
  title={Pandora: Towards general world model with natural language actions and video states},
  author={Xiang, Jiannan and Liu, Guangyi and Gu, Yi and Gao, Qiyue and Ning, Yuting and Zha, Yuheng and Feng, Zeyu and Tao, Tianhua and Hao, Shibo and Shi, Yemin and others},
  journal={arXiv preprint arXiv:2406.09455},
  year={2024}
}

@article{opensoraplan,
  title={Open-sora plan: Open-source large video generation model},
  author={Lin, Bin and Ge, Yunyang and Cheng, Xinhua and Li, Zongjian and Zhu, Bin and Wang, Shaodong and He, Xianyi and Ye, Yang and Yuan, Shenghai and Chen, Liuhan and others},
  journal={arXiv preprint arXiv:2412.00131},
  year={2024}
}

@inproceedings{cogvideox,
  title={Cogvideox: Text-to-video diffusion models with an expert transformer},
  author={Yang, Zhuoyi and Teng, Jiayan and Zheng, Wendi and Ding, Ming and Huang, Shiyu and Xu, Jiazheng and Yang, Yuanming and Hong, Wenyi and Zhang, Xiaohan and Feng, Guanyu and others},
  booktitle={International Conference on Learning Representations},
  volume={2025},
  pages={83048--83077},
  year={2025}
}

@misc{mochi,
  title        = {{Mochi 1}},
  author       = {{Genmo Team}},
  year         = {2024},
  howpublished = {\url{https://github.com/genmoai/models}},
  note         = {Accessed: April 29, 2026}
}

@misc{luma,
  title        = {{Luma Dream Machine}: {AI} Video Generator},
  author       = {{Luma AI}},
  year         = {2024},
  howpublished = {\url{https://dream-machine.lumalabs.ai/}},
  note         = {Accessed: April 29, 2026}
}

@misc{runway,
  title        = {Introducing {Gen-3 Alpha}},
  author       = {{Runway ML}},
  year         = {2024},
  howpublished = {\url{https://runwayml.com/research/introducing-gen-3-alpha}},
  note         = {Accessed: April 29, 2026}
}

@article{qwenimage,
  title={Qwen-image technical report},
  author={Wu, Chenfei and Li, Jiahao and Zhou, Jingren and Lin, Junyang and Gao, Kaiyuan and Yan, Kun and Yin, Sheng-ming and Bai, Shuai and Xu, Xiao and Chen, Yilei and others},
  journal={arXiv preprint arXiv:2508.02324},
  year={2025}
}

@article{inter4k,
  title={Adapool: Exponential adaptive pooling for information-retaining downsampling},
  author={Stergiou, Alexandros and Poppe, Ronald},
  journal={IEEE Transactions on Image Processing},
  volume={32},
  pages={251--266},
  year={2022},
  publisher={IEEE}
}

@article{yi2021robustness,
  title={Benchmarking the robustness of spatial-temporal models against corruptions},
  author={Yi, Chenyu and Yang, Siyuan and Li, Haoliang and Tan, Yap-peng and Kot, Alex},
  journal={arXiv preprint arXiv:2110.06513},
  year={2021}
}

@inproceedings{tempocontrol,
  title={Tempocontrol: Temporal attention guidance for text-to-video models},
  author={Schiber, Shira and Lindenbaum, Ofir and Schwartz, Idan},
  booktitle={Proceedings of the IEEE/CVF Conference on Computer Vision and Pattern Recognition},
  pages={36670--36679},
  year={2026}
}

@article{adaworld,
  title={Adaworld: Learning adaptable world models with latent actions},
  author={Gao, Shenyuan and Zhou, Siyuan and Du, Yilun and Zhang, Jun and Gan, Chuang},
  journal={arXiv preprint arXiv:2503.18938},
  year={2025}
}

@inproceedings{arlon,
  title={Arlon: Boosting diffusion transformers with autoregressive models for long video generation},
  author={Li, Zongyi and Hu, Shujie and Liu, Shujie and Zhou, Long and Choi, Jeongsoo and Meng, Lingwei and Guo, Xun and Li, Jinyu and Ling, Hefei and Wei, Furu},
  booktitle={International Conference on Learning Representations},
  volume={2025},
  pages={90986--91004},
  year={2025}
}

@article{videomar,
  title={Videomar: Autoregressive video generation with continuous tokens},
  author={Yu, Hu and Gong, Biao and Yuan, Hangjie and Zheng, DanDan and Chai, Weilong and Chen, Jingdong and Zheng, Kecheng and Zhao, Feng},
  journal={Advances in neural information processing systems},
  volume={38},
  pages={56928--56958},
  year={2026}
}

@article{anyv2v,
  title={Anyv2v: A tuning-free framework for any video-to-video editing tasks},
  author={Ku, Max and Wei, Cong and Ren, Weiming and Yang, Harry and Chen, Wenhu},
  journal={arXiv preprint arXiv:2403.14468},
  year={2024}
}

@article{worldbench,
  title={WorldBench: Benchmarking Physical Understanding of World Models by Isolating Physics Concepts},
  author={Upadhyay, Rishi and Zhang, Howard and Solomon, Jim and Agrawal, Ayush and Ba, Yunhao and Wong, Alex and de Melo, Celso M and Kadambi, Achuta},
  journal={arXiv preprint arXiv:2601.21282},
  year={2026}
}

@article{opsd,
  title={Self-distilled reasoner: On-policy self-distillation for large language models},
  author={Zhao, Siyan and Xie, Zhihui and Liu, Mengchen and Huang, Jing and Pang, Guan and Chen, Feiyu and Grover, Aditya},
  journal={arXiv preprint arXiv:2601.18734},
  year={2026}
}

@inproceedings{instructpix2pix,
  title={Instructpix2pix: Learning to follow image editing instructions},
  author={Brooks, Tim and Holynski, Aleksander and Efros, Alexei A},
  booktitle={2023 IEEE/CVF Conference on Computer Vision and Pattern Recognition (CVPR)},
  pages={18392--18402},
  year={2023},
  organization={IEEE}
}

@article{videodirectorgpt,
  title={Videodirectorgpt: Consistent multi-scene video generation via llm-guided planning},
  author={Lin, Han and Zala, Abhay and Cho, Jaemin and Bansal, Mohit},
  journal={arXiv preprint arXiv:2309.15091},
  year={2023}
}

@article{freebloom,
  title={Free-bloom: Zero-shot text-to-video generator with llm director and ldm animator},
  author={Huang, Hanzhuo and Feng, Yufan and Shi, Cheng and Xu, Lan and Yu, Jingyi and Yang, Sibei},
  journal={Advances in Neural Information Processing Systems},
  volume={36},
  pages={26135--26158},
  year={2023}
}

@inproceedings{vlogger,
  title={Vlogger: Make your dream a vlog},
  author={Zhuang, Shaobin and Li, Kunchang and Chen, Xinyuan and Wang, Yaohui and Liu, Ziwei and Qiao, Yu and Wang, Yali},
  booktitle={2024 IEEE/CVF Conference on Computer Vision and Pattern Recognition (CVPR)},
  pages={8806--8817},
  year={2024},
  organization={IEEE}
}

@article{mora,
  title={Mora: Enabling generalist video generation via a multi-agent framework},
  author={Yuan, Zhengqing and Liu, Yixin and Cao, Yihan and Sun, Weixiang and Jia, Haolong and Chen, Ruoxi and Li, Zhaoxu and Lin, Bin and Yuan, Li and He, Lifang and others},
  journal={arXiv preprint arXiv:2403.13248},
  year={2024}
}

@article{vgot,
  title={Videogen-of-thought: Step-by-step generating multi-shot video with minimal manual intervention},
  author={Zheng, Mingzhe and Xu, Yongqi and Huang, Haojian and Ma, Xuran and Liu, Yexin and Shu, Wenjie and Pang, Yatian and Tang, Feilong and Chen, Qifeng and Yang, Harry and others},
  journal={arXiv preprint arXiv:2412.02259},
  year={2024}
}

@article{movieagent,
  title={Automated movie generation via multi-agent cot planning},
  author={Wu, Weijia and Zhu, Zeyu and Shou, Mike Zheng},
  journal={arXiv preprint arXiv:2503.07314},
  year={2025}
}

@article{dreamfactory,
  title={Dreamfactory: Pioneering multi-scene long video generation with a multi-agent framework},
  author={Xie, Zhifei and Tang, Daniel and Tan, Dingwei and Klein, Jacques and Bissyand, Tegawend F and Ezzini, Saad},
  journal={arXiv preprint arXiv:2408.11788},
  year={2024}
}

@inproceedings{genmac,
  title={Genmac: compositional text-to-video generation with multi-agent collaboration},
  author={Huang, Kaiyi and Huang, Yukun and Ning, Xuefei and Lin, Zinan and Wang, Yu and Liu, Xihui},
  booktitle={Proceedings of the AAAI Conference on Artificial Intelligence},
  volume={40},
  number={7},
  pages={5049--5057},
  year={2026}
}

@article{deltaworld,
  title={A frame is worth one token: Efficient generative world modeling with delta tokens},
  author={Kerssies, Tommie and Berton, Gabriele and He, Ju and Yu, Qihang and Ma, Wufei and de Geus, Daan and Dubbelman, Gijs and Chen, Liang-Chieh},
  journal={arXiv preprint arXiv:2604.04913},
  year={2026}
}

@article{deltawam,
  title={DeltaWAM: Change-Centric Visual Foresight via Delta Tokens for an Efficient World-Action Model},
  author={Jiang, Tianyun and Bao, Wenrui and Xu, Bingxin and Tian, Yu and Shang, Yuzhang},
  journal={arXiv preprint arXiv:2609.33177},
  year={2026}
}

@article{videoworld2,
  title={Videoworld 2: Learning transferable knowledge from real-world videos},
  author={Ren, Zhongwei and Wei, Yunchao and Yu, Xiao and Luo, Guixun and Zhao, Yao and Kang, Bingyi and Feng, Jiashi and Jin, Xiaojie},
  journal={arXiv preprint arXiv:2602.10102},
  year={2026}
}

@inproceedings{landiff,
  title={The best of both worlds: Integrating language models and diffusion models for video generation},
  author={Yin, Aoxiong and Tan, Xu and Shen, Kai and Leng, Yichong and Zhou, Xinyu and Li, Juncheng and Tang, Siliang},
  booktitle={2025 IEEE/CVF International Conference on Computer Vision (ICCV)},
  pages={01--12},
  year={2025},
  organization={IEEE}
}

@article{planx,
  title={Plan-X: Instruct video generation via semantic planning},
  author={Huang, Lun and Xie, You and Xu, Hongyi and Gu, Tianpei and Zhang, Chenxu and Song, Guoxian and Li, Zenan and Zhao, Xiaochen and Luo, Linjie and Sapiro, Guillermo},
  journal={arXiv preprint arXiv:2511.17986},
  year={2025}
}

@article{bividgen,
  title={Beyond Text Conditioning: A Systematic Study of MLLM-DiT Fusion for Video Generation},
  author={Ding, Yanbo and Fan, Yijia and Shan, Caihua and Yang, Yifan and Shen, Yifei and Wang, Weijie and Hu, Xirui and Li, Dongsheng and Qiu, Lili and Yang, Yuqing and others},
  journal={arXiv preprint arXiv:2608.14043},
  year={2026}
}

@article{vtok,
  title={VTok: A Unified Video Tokenizer with Decoupled Spatial-Temporal Latents},
  author={Wang, Feng and Shi, Yichun and Yang, Ceyuan and Guo, Qiushan and Sun, Jingxiang and Yuille, Alan and Wang, Peng},
  journal={arXiv preprint arXiv:2602.04202},
  year={2026}
}

@article{pan,
  title={Pan: A world model for general, interactable, and long-horizon world simulation},
  author={Xiang, Jiannan and Gu, Yi and Liu, Zihan and Feng, Zeyu and Gao, Qiyue and Hu, Yiyan and Huang, Benhao and Liu, Guangyi and Yang, Yichi and Zhou, Kun and others},
  journal={arXiv preprint arXiv:2511.09057},
  year={2025}
}

@article{bernini,
  title={Bernini: Latent Semantic Planning for Video Diffusion},
  author={Team, Bernini and Liu, Chenchen and Chen, Junyi and Li, Lei and Chi, Lu and Sun, Mingzhen and Li, Zhuoying and Fu, Yi and Guo, Ruoyu and Wu, Yiheng and others},
  journal={arXiv preprint arXiv:2605.22344},
  year={2026}
}

@article{phyworld,
  title={How far is video generation from world model: A physical law perspective},
  author={Kang, Bingyi and Yue, Yang and Lu, Rui and Lin, Zhijie and Zhao, Yang and Wang, Kaixin and Huang, Gao and Feng, Jiashi},
  journal={arXiv preprint arXiv:2411.02385},
  year={2024}
}

@inproceedings{lapa,
  title={Latent action pretraining from videos},
  author={Ye, Seonghyeon and Jang, Joel and Jeon, Byeongguk and Joo, Se June and Yang, Jianwei and Peng, Baolin and Mandlekar, Ajay and Tan, Reuben and Chao, Yu-Wei and Lin, Bill Yuchen and others},
  booktitle={International Conference on Learning Representations},
  volume={2025},
  pages={28213--28239},
  year={2025}
}

@article{univla,
  title={Univla: Learning to act anywhere with task-centric latent actions},
  author={Bu, Qingwen and Yang, Yanting and Cai, Jisong and Gao, Shenyuan and Ren, Guanghui and Yao, Maoqing and Luo, Ping and Li, Hongyang},
  journal={arXiv preprint arXiv:2505.06111},
  year={2025}
}

@inproceedings{moto,
  title={Moto: Latent motion token as the bridging language for learning robot manipulation from videos},
  author={Chen, Yi and Ge, Yuying and Tang, Weiliang and Li, Yizhuo and Ge, Yixiao and Ding, Mingyu and Shan, Ying and Liu, Xihui},
  booktitle={2025 IEEE/CVF International Conference on Computer Vision (ICCV)},
  pages={19752--19763},
  year={2025},
  organization={IEEE}
}

@inproceedings{genie,
  title={Genie: Generative interactive environments},
  author={Bruce, Jake and Dennis, Michael D and Edwards, Ashley and Parker-Holder, Jack and Shi, Yuge and Hughes, Edward and Lai, Matthew and Mavalankar, Aditi and Steigerwald, Richie and Apps, Chris and others},
  booktitle={Forty-first international conference on machine learning},
  year={2024}
}

@article{hunyuan,
  title={Hunyuanvideo: A systematic framework for large video generative models},
  author={Kong, Weijie and Tian, Qi and Zhang, Zijian and Min, Rox and Dai, Zuozhuo and Zhou, Jin and Xiong, Jiangfeng and Li, Xin and Wu, Bo and Zhang, Jianwei and others},
  journal={arXiv preprint arXiv:2412.03603},
  year={2024}
}
}

\appendix
% Appendix: title block, overview with an appendix-only table of contents,
% and figure/table references in red bold.
\twocolumn[
  \centering
  \Large
  \textbf{\thetitle}\\
  Supplementary Material\\
  \vspace{1.0em}
]
\suppressfloats[t]

\makeatletter
% table of contents listing only the appendix sections, with dotted leaders
\newcommand{\appendixtocstart}{%
  \renewcommand*{\l@section}[2]{\par\addvspace{0.6em}\noindent{\bfseries\color{cvprblue}\setlength{\@tempdima}{1.8em}##1}\nobreak\hfill{\bfseries ##2}\par}%
  \renewcommand*{\l@subsection}[2]{\par\noindent\hspace*{1.4em}{\color{cvprblue}\setlength{\@tempdima}{2.4em}##1}\nobreak\leaders\hbox{$\m@th\mkern 4.5mu.\mkern 4.5mu$}\hfill\nobreak ##2\par}}
\newcommand{\appendixcontents}{\begingroup\let\l@section\@gobbletwo\let\l@subsection\@gobbletwo\let\l@subsubsection\@gobbletwo\@starttoc{toc}\endgroup}
\makeatother
\addtocontents{toc}{\protect\appendixtocstart}

% in-text references to figures and tables in the appendix: full names, red and bold
\newcommand{\redref}[1]{\textbf{\textcolor{red}{#1}}}
\newcommand{\refnum}[3]{#2\textbf{#1}#3}
\crefformat{figure}{\redref{Figure}~\refnum{#1}{#2}{#3}}
\Crefformat{figure}{\redref{Figure}~\refnum{#1}{#2}{#3}}
\crefmultiformat{figure}{\redref{Figures}~\refnum{#1}{#2}{#3}}{ and~\refnum{#1}{#2}{#3}}{, \refnum{#1}{#2}{#3}}{, and~\refnum{#1}{#2}{#3}}
\Crefmultiformat{figure}{\redref{Figures}~\refnum{#1}{#2}{#3}}{ and~\refnum{#1}{#2}{#3}}{, \refnum{#1}{#2}{#3}}{, and~\refnum{#1}{#2}{#3}}
\crefrangeformat{figure}{\redref{Figures}~\refnum{#1}{#3}{#4} to~\refnum{#2}{#5}{#6}}
\Crefrangeformat{figure}{\redref{Figures}~\refnum{#1}{#3}{#4} to~\refnum{#2}{#5}{#6}}
\crefformat{table}{\redref{Table}~\refnum{#1}{#2}{#3}}
\Crefformat{table}{\redref{Table}~\refnum{#1}{#2}{#3}}
\crefmultiformat{table}{\redref{Tables}~\refnum{#1}{#2}{#3}}{ and~\refnum{#1}{#2}{#3}}{, \refnum{#1}{#2}{#3}}{, and~\refnum{#1}{#2}{#3}}
\Crefmultiformat{table}{\redref{Tables}~\refnum{#1}{#2}{#3}}{ and~\refnum{#1}{#2}{#3}}{, \refnum{#1}{#2}{#3}}{, and~\refnum{#1}{#2}{#3}}
\crefrangeformat{table}{\redref{Tables}~\refnum{#1}{#3}{#4} to~\refnum{#2}{#5}{#6}}
\Crefrangeformat{table}{\redref{Tables}~\refnum{#1}{#3}{#4} to~\refnum{#2}{#5}{#6}}

\noindent\textbf{Overview}
In this supplementary material, we present the following content.

\medskip
\appendixcontents
\medskip

%==========================================================
\section{Implementation Details}
\label{sec:supp_impl}

\subsection{Tokenizer}
\label{sec:supp_tok}

\noindent\textbf{Encoder.}
The encoder is the frozen SigLIP2-g/16 vision encoder of X-Omni~\cite{xomni}.
For a 480p frame, it yields a grid of $\tfrac{H}{16}\times\tfrac{W}{16}$ patches, \ie, 1,404--1,680 patches across our aspect-ratio buckets, and we take the 1536-d features after the last of its 40 blocks and before the final normalization.
The features are standardized per channel with the mean and standard deviation estimated on training frames.

\noindent\textbf{Quantizer.}
A Q-Former with 32 learnable queries cross-attends to the features of the two keyframes of a chunk, which are 8 frames apart at 24\,fps.
Two linear layers project each of the 32 output vectors from 1536 to 48 and then to 6 dimensions, and FSQ quantizes the 6 dimensions to $(8,5,5,5,5,5)$ levels, giving a vocabulary of 25,000 delta tokens.
Each chunk is quantized independently, so a five-second clip yields 15 chunks and 480 tokens.

\noindent\textbf{Decoder.}
The decoder consists of 24 transformer blocks initialized from the last 24 blocks of SigLIP2-g.
Its input concatenates the encoder features of the first keyframe, with the SigLIP position embeddings added, and the 32 delta tokens, embedded by an MLP with learnable position embeddings.
Through the frozen projection and codebook of X-Omni, its outputs at the positions of the first keyframe give a distribution over the 16,384 visual tokens at each patch, which is trained to match that of the second keyframe by the KL objective in \cref{eq:loss}.
The Q-Former, the projections, and the decoder have 709M trainable parameters in total.
The decoder is used only for training, and at inference only the encoder and the quantizer extract the delta tokens of the input video.

\noindent\textbf{Training.}
The tokenizer is trained with the loss in \cref{eq:loss}, and \cref{tab:train_cfg} lists its training hyper-parameters.

\begin{table}[t]
\centering
\caption{Detailed training hyper-parameters of each stage.}
\label{tab:train_cfg}
\fontsize{6.2}{7.2}\selectfont\renewcommand{\arraystretch}{1.15}
\setlength{\tabcolsep}{1.5pt}
\begin{tabular*}{\linewidth}{@{\extracolsep{\fill}}lccccc@{}}
\toprule
\textbf{Hyper-parameters} & \textbf{Tokenizer} & \textbf{VLM-V2V} & \textbf{VLM-TD} & \textbf{VLM-TI2V} & \textbf{Diffusion} \\
\hline
Backbone init. & SigLIP2-g & Qwen3-VL-8B & VLM-V2V & VLM-TD & Wan2.2-5B \\
Optimizer & AdamW & AdamW & AdamW & AdamW & AdamW \\
Optimizer param. & \multicolumn{5}{c}{$\beta_1 = 0.9, \beta_2 = 0.99, \epsilon = 1\mathrm{e}{-}8$} \\
LR & 1e-4 & 1e-4 & 1e-6 & 1e-4 & 1e-4 \\
LR scheduler & Constant & Constant & Constant & Constant & Constant \\
Warmup steps & 500 & 1,000 & 50 & 1,000 & 1,000 \\
Weight decay & 0.01 & 0.0 & 0.0 & 0.0 & 0.0 \\
Gradient clipping & 1.0 & 1.0 & 1.0 & 1.0 & 1.0 \\
EMA & 0.9995 & 0.999 & -- & 0.999 & 0.999 \\
Numerical precision & float16 & bfloat16 & bfloat16 & bfloat16 & bfloat16 \\
Training clips & 5.67M & 5.67M & 2.03M & 2.03M & 2.03M \\
Batch size & 2,048 frame pairs & 768 clips & 384 clips & 768 clips & 64 clips \\
Training steps & 100K & 20K & 7K & 8K & 56K \\
Resource usage & 128 A800 & 64 A800 & 32 A800 & 64 A800 & 64 A800 \\
Framework & DDP & ZeRO-2 & ZeRO-2 & ZeRO-2 & DDP \\
\bottomrule
\end{tabular*}
\end{table}

\subsection{Autoregressive Vision-Language Model}
\label{sec:supp_ar}

\noindent\textbf{Architecture.}
The VLM backbone is initialized from Qwen3-VL-8B, and its vocabulary is expanded with the 25,000 delta tokens, whose input and output embeddings are initialized from the MLP that embeds the tokens in the tokenizer decoder.
We add 32 slot embeddings, initialized from the learnable position embeddings of the tokens in the tokenizer decoder, to mark the position of each token within its chunk.
The vision encoder is replaced by the frozen SigLIP2-g of the tokenizer, followed by a trainable projection.
All parameters except SigLIP2-g are trained in all stages.

\noindent\textbf{Training.}
Each sample is a chat in which the user turn holds the first frame and the prompt ``Text caption: \{caption\}. Predict the motion token sequence.'', where the caption is empty for unlabeled videos, and the assistant turn holds the delta tokens between two new special tokens, \texttt{<bom>} and \texttt{<eom>}.
The AR model is trained in three stages (\cref{sec:ar}): V2V pretraining, trajectory distillation (TD), and TI2V fine-tuning.
The V2V data contain clips of up to five seconds, about 1.08M of which are shorter than five seconds.
TD and TI2V fine-tuning use 2.03M five-second clips with high dynamics, and TD drops the captions.
V2V pretraining excludes \texttt{<bom>} and the first 96 tokens from the loss, and TD starts from the V2V checkpoint, which also serves as the frozen teacher.
\Cref{tab:train_cfg} lists the hyper-parameters of each stage.
\Cref{alg:td} summarizes trajectory distillation (\cref{sec:ar} and \cref{eq:td}), and \cref{sec:supp_td_real} analyzes its effect on WorldBench.
In a five-second clip, the student starts from the keyframe at 2\,s, $x$ contains the ground-truth tokens of the next second (96 tokens), and the student samples the tokens $\hat y$ of the last 2\,s (192 tokens).
The teacher additionally sees $y^*$, the ground-truth tokens from 0\,s to 2\,s (192 tokens).
Both distributions are computed by a softmax over the delta tokens only, and gradients flow only through the student.

\begin{algorithm}[t]
\caption{Trajectory Distillation}
\label{alg:td}
\begin{algorithmic}[1]
\Require dataset $\mathcal{S}=\{(x_i,y_i^*)\}_{i=1}^{N}$; pretrained AR model $p_\theta$; divergence $D$ (forward KL)
\State Let $p_S(\cdot\mid x)$ and $p_T(\cdot\mid x,y^*)$ be the same model $p_\theta$ given ground-truth tokens of different lengths, with $p_T$ frozen
\While{not converged}
  \State Sample a minibatch $\mathcal{B} \subset \mathcal{S}$
  \ForAll{$(x,y^*) \in \mathcal{B}$}
    \State Sample an on-policy rollout $\hat y \sim p_S(\cdot \mid x)$
    \State Token-wise divergence along the student rollout:
    \Statex $\ell(x,y^*) \gets \tfrac{1}{|\hat y|}\sum_n D\bigl(p_T(\cdot|x,y^*,\hat y_{<n}) \| p_S(\cdot|x,\hat y_{<n})\bigr)$
  \EndFor
  \State $\loss_{\text{TD}}(\theta) \gets \tfrac{1}{|\mathcal{B}|}\sum_{(x,y^*)\in\mathcal{B}} \ell(x,y^*)$; update $\theta$
\EndWhile
\end{algorithmic}
\end{algorithm}

\noindent\textbf{Inference.}
To generate videos longer than the five-second training clips, the AR model keeps sampling delta tokens chunk by chunk, with all previous tokens and the first frame kept in its context and no limit on the context length.
For VBench-Long~\cite{vbenchpp}, we use its official prompts and, following ARLON~\cite{arlon}, generate 25-second videos at 24\,fps with 601 frames.
For WorldBench~\cite{worldbench}, Physics-IQ Verified~\cite{physicsiq}, PhyGround~\cite{phyground}, WorldModelBench~\cite{worldmodelbench}, VBench-I2V~\cite{vbenchpp}, and TempoControl~\cite{tempocontrol}, we generate five-second videos at 24\,fps with 121 frames.
On WorldBench, the 33-frame results are evaluated on the first 33 frames of the 121-frame continuations.
We sample the delta tokens with a temperature of 1. The top-$k$ of 50 in \cref{sec:ar} applies to videos of up to five seconds, while long videos use top-$k$ sampling with $k\!=\!4{,}096$.

\subsection{Diffusion Renderer}
\label{sec:supp_renderer}

\noindent\textbf{Architecture.}
The renderer is Wan2.2-TI2V-5B with LoRA of rank 32, and the rest of the diffusion transformer stays frozen.
Three conditions enter through cross-attention: the umT5 embedding of the caption as the text condition $c_T$, and the SigLIP2-g features of the first frame together with the embeddings of the 480 delta tokens (15 chunks $\times$ 32) as the visual condition $c_I$ (\cref{sec:renderer}).
The embeddings of the delta tokens and of the SigLIP2-g features are initialized from the corresponding layers of the tokenizer decoder, the delta tokens carry learnable position embeddings within a chunk, and new learnable embeddings mark the 15 chunks.
They are trained with the LoRA weights.

\noindent\textbf{Training.}
The renderer is trained on the 2.03M captioned five-second clips and their delta tokens, independently of the AR model.
To enable classifier-free guidance on both conditions (\cref{eq:cfg}), the text condition $c_T$ alone, the visual condition $c_I$ alone, or both are randomly dropped during training, each with a probability of 10\%.

\noindent\textbf{Inference.}
We use classifier-free guidance scales of $s_I\!=\!3$ and $s_T\!=\!5$ in \cref{eq:cfg}.
For long videos, the renderer generates the video segment by segment, each segment starting from the last frame of the previous one.
When the AR model continues a video beyond what the text prompt describes, the continued segment is rendered without text.

%==========================================================
\section{Training Data}
\label{sec:supp_data}

\Cref{tab:data} lists the number of clips from each dataset.
All training data come from open-source datasets.
The V2V data contain 5.67M clips of up to five seconds (121 frames), about 1.08M of which are shorter than five seconds.
Trajectory distillation, TI2V fine-tuning, and the renderer share 2.03M five-second clips with high dynamics.

\begin{table}[t]
\centering
\caption{Training data. V2V pretraining uses clips of up to five seconds, while trajectory distillation (TD), TI2V fine-tuning, and the renderer use five-second clips with high dynamics.}
\label{tab:data}
\fontsize{6.5}{7.5}\selectfont\renewcommand{\arraystretch}{1.15}
\setlength{\tabcolsep}{3pt}
\begin{tabular*}{\linewidth}{@{\extracolsep{\fill}}llcc@{}}
\toprule
\textbf{Dataset} & \textbf{Type} & \textbf{Clips (V2V)} & \textbf{Clips (TD / TI2V / Diffusion)} \\
\hline
HOIGen-1M~\cite{hoigen} & Real & 1,177K & 475K \\
Open-Sora Plan~\cite{opensoraplan} & Real & 801K & -- \\
OpenVid-1M~\cite{openvid} & Real & 700K & 313K \\
VIDGEN-1M~\cite{vidgen} & Real & 686K & 307K \\
Sekai-walking~\cite{sekai} & Real & 581K & 102K \\
EgoDex~\cite{egodex} & Real & 322K & 52K \\
LLaVA-OneVision-2~\cite{llavaov2} & Real & 322K & 176K \\
Kinetics-700~\cite{kinetics700} & Real & 115K & 56K \\
Moments in Time~\cite{momentsintime} & Real & 80K & -- \\
ActivityNet~\cite{activitynet} & Real & 79K & 47K \\
BAAI-Physics & Real & 77K & -- \\
UltraVideo~\cite{ultravideo} & Real & 64K & 36K \\
EgoDemo & Real & 59K & 45K \\
Kinetics-400~\cite{kinetics400} & Real & 54K & 35K \\
WISA-80K~\cite{wisa} & Real & 46K & 12K \\
Galaxea Open-World~\cite{galaxea} & Real & 41K & 39K \\
EPIC-KITCHENS~\cite{epickitchens} & Real & 20K & 19K \\
Oops!~\cite{oops} & Real & 10K & 6K \\
Sekai-drone~\cite{sekai} & Real & 8K & 1K \\
MotionBench~\cite{motionbench} & Real & 7K & 4K \\
MultiSports~\cite{multisports} & Real & 7K & 883 \\
TemporalBench~\cite{temporalbench} & Real & 4K & 1K \\
\hline
Cosmos3~\cite{cosmos3} & Simulation & 251K & 251K \\
Phyco~\cite{phyco} & Simulation & 68K & -- \\
ComPhy~\cite{comphy} & Simulation & 42K & 9K \\
Physion~\cite{physion} & Simulation & 20K & 20K \\
Physion++~\cite{physionpp} & Simulation & 14K & 14K \\
CLEVRER~\cite{clevrer} & Simulation & 10K & 10K \\
Kubric MOVi~\cite{kubric} & Simulation & 6K & -- \\
\hline
Total & -- & 5.67M & 2.03M \\
\bottomrule
\end{tabular*}
\end{table}

%==========================================================
\section{Analysis}
\label{sec:supp_analysis}

\subsection{Trajectory Distillation on WorldBench}
\label{sec:supp_td_real}

\begin{figure*}[p]
\centering
\includegraphics[width=\linewidth,height=0.86\textheight,keepaspectratio]{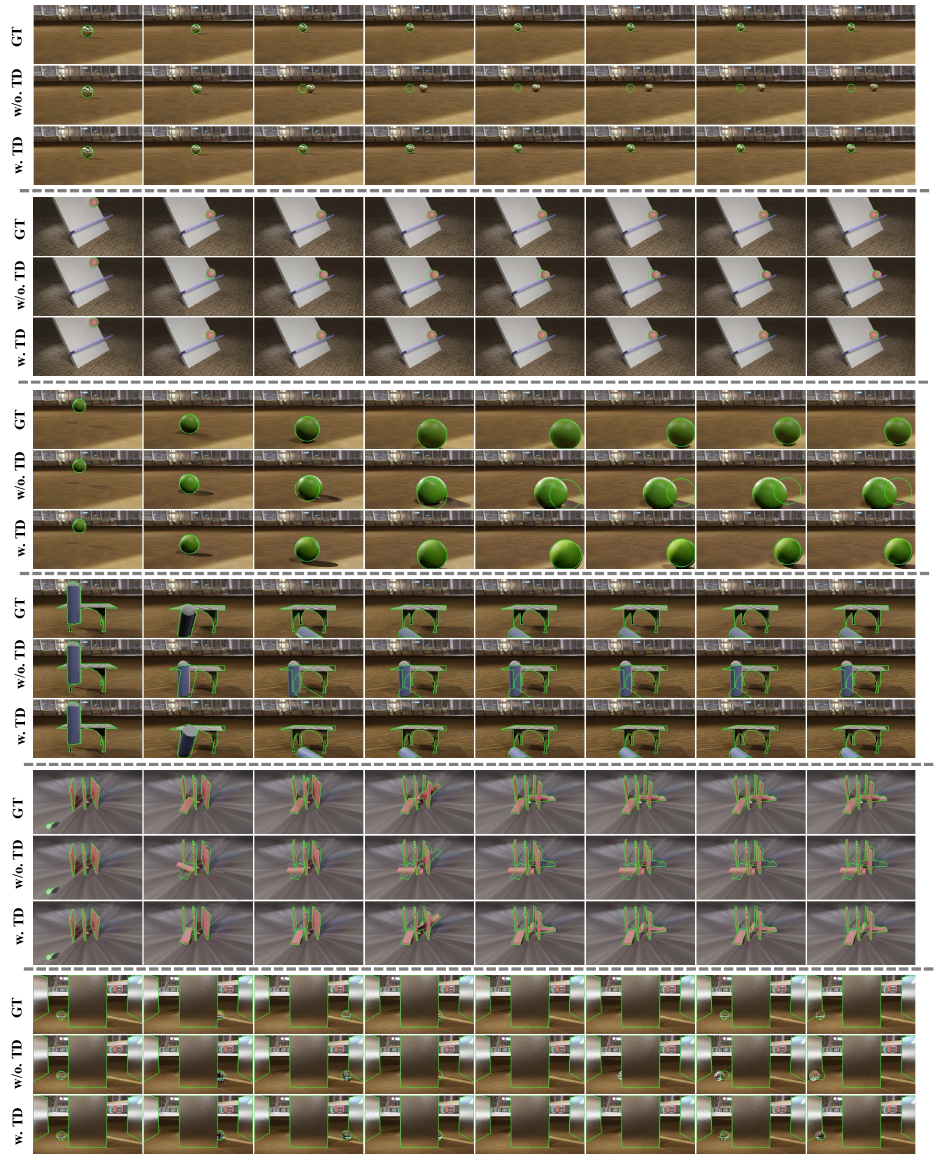}
\caption{V2V continuations on WorldBench before and after trajectory distillation.}
\label{fig:supp_worldbench}
\end{figure*}

In \cref{fig:supp_worldbench}, we compare the V2V continuations on the simulated scenes of WorldBench~\cite{worldbench} before and after trajectory distillation.
The green outlines mark the ground-truth foreground masks, on which WorldBench computes the mIoU.
Without trajectory distillation, errors accumulate over chunks, and objects drift from their ground-truth trajectories and violate physical laws.
With trajectory distillation, the continuations stay close to the ground truth, consistent with the gains in \cref{tab:worldbench}, since the student learns to correct its errors from its own rollouts under a teacher that sees a longer ground-truth history.

\subsection{Tokenizer Training Objective}
\label{sec:supp_tok_obj}

\begin{figure*}[t]
\centering
\includegraphics[width=\linewidth,height=0.86\textheight,keepaspectratio]{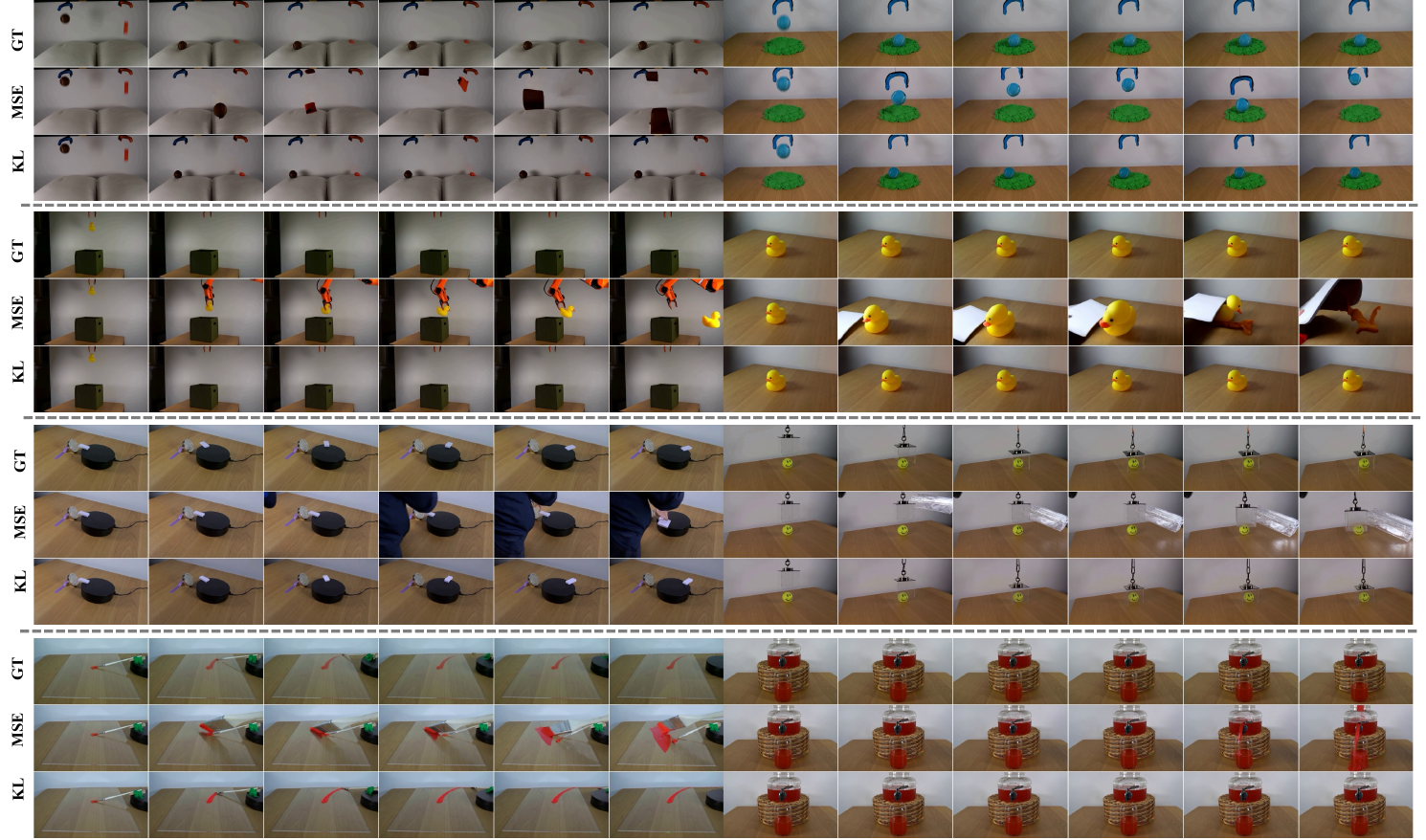}
\caption{TI2V generation on Physics-IQ Verified with the tokenizers trained by MSE and KL.}
\label{fig:supp_tok_physiq}
\end{figure*}

\begin{figure*}[t]
\centering
\includegraphics[width=\linewidth,height=0.86\textheight,keepaspectratio]{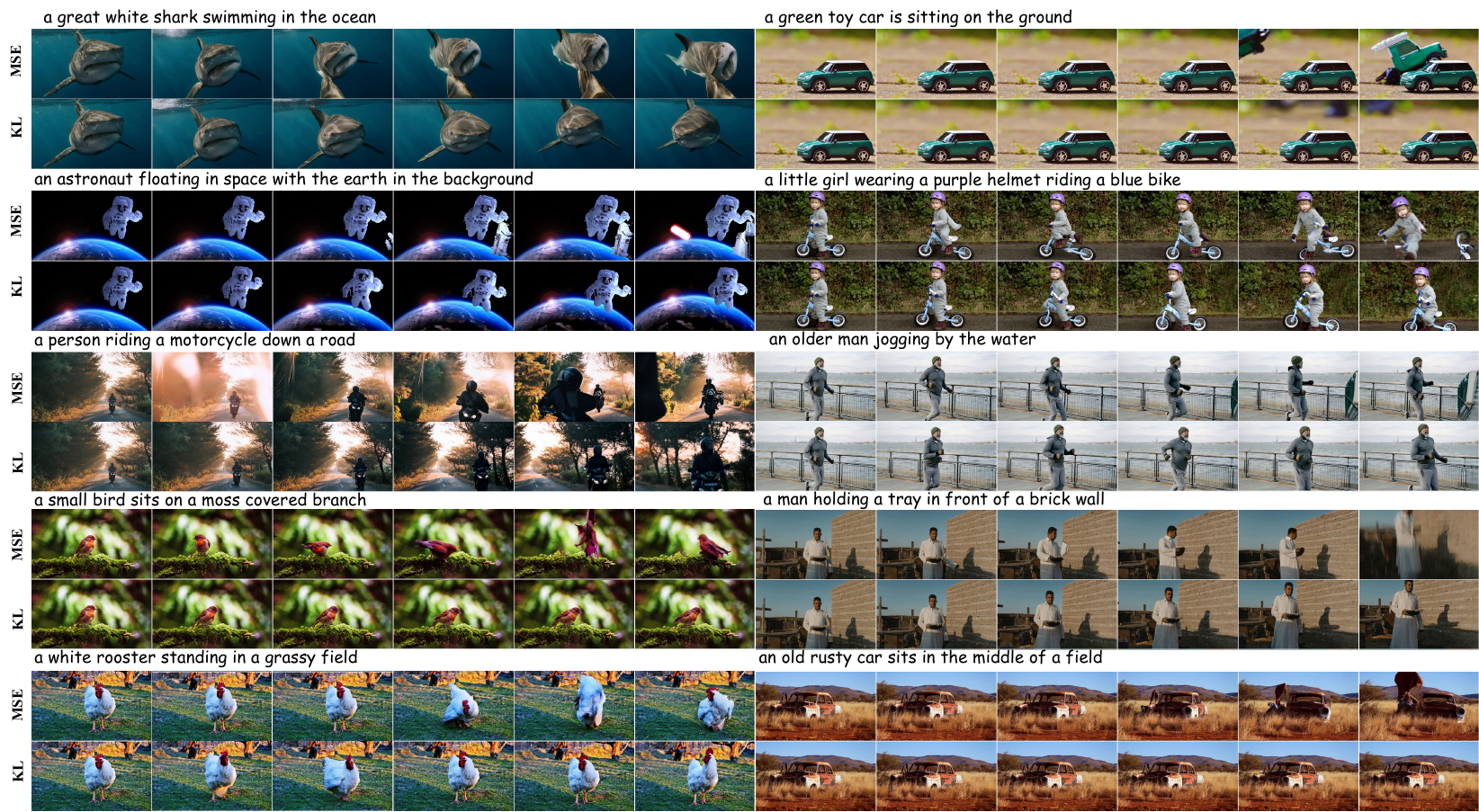}
\caption{TI2V generation on VBench-I2V with the tokenizers trained by MSE and KL.}
\label{fig:supp_tok_vbench}
\end{figure*}

\Cref{tab:abl_tok} measures the fraction of delta tokens that flip under per-frame perturbations on Inter4K, and \cref{tab:pert_psnr} lists the parameter and PSNR of each perturbation level, where the perturbations follow~\cite{yi2021robustness} and the PSNR is measured on the 16 keyframes of each five-second video.
Even at Level 1, where the PSNR stays above 41\,dB, the MSE tokenizer flips the majority of delta tokens, so its tokens mix semantic changes with noise in the features.
We examine how this affects the AR model.
To isolate the effect of the training objective, we start from \ours$_{\text{w/o.\ V2V}}$ in \cref{tab:abl_v2v}, which is trained directly for TI2V from Qwen3-VL-8B with the KL tokenizer, and train another AR model in the same way with the MSE tokenizer for the same number of steps.
\Cref{fig:supp_tok_physiq,fig:supp_tok_vbench} show results on Physics-IQ Verified~\cite{physicsiq} and VBench-I2V~\cite{vbenchpp}.
With the MSE tokenizer, frame pairs with small semantic changes still yield many different tokens, so the AR model can hardly tell noise from semantic change.
It therefore generates motions that violate the dynamics and fails to preserve subtle motions.
With the KL tokenizer, the tokens follow changes in semantics, and the AR model keeps both subtle and large motions consistent with the scene.

\begin{table}[t]
\centering
\caption{Perturbation levels on Inter4K~\cite{inter4k}. Each cell lists the parameter / PSNR (dB) with respect to the clean video. CRF: H.264 constant rate factor. $c$: photon count of shot noise, $x' = \mathrm{Poisson}(c\,x)/c$. $\delta$: maximum brightness shift, adding $\mathcal{U}(0,\delta)$ to the HSV value channel. $f$: minimum contrast factor, $x' = (x-\mu)\,s + \mu$ with $s \sim \mathcal{U}(f,1)$. Each frame is perturbed independently, following~\cite{yi2021robustness}.}
\label{tab:pert_psnr}
\fontsize{6.5}{7.5}\selectfont\renewcommand{\arraystretch}{1.15}
\setlength{\tabcolsep}{3pt}
\begin{tabular*}{\linewidth}{@{\extracolsep{\fill}}lcccc@{}}
\toprule
\multirow{2}{*}{\textbf{Perturbation}} & \textbf{Level 1} & \textbf{Level 2} & \textbf{Level 3} & \textbf{Level 4} \\
 & Param.\ / PSNR & Param.\ / PSNR & Param.\ / PSNR & Param.\ / PSNR \\
\hline
H.264 (CRF) & 12 / 41.7 & 18 / 37.8 & 23 / 34.7 & 30 / 30.5 \\
Shot noise ($c$) & 5000 / 41.4 & 2500 / 38.6 & 1000 / 34.7 & 400 / 30.8 \\
Brightness ($\delta$) & 0.02 / 42.0 & 0.03 / 38.0 & 0.05 / 33.3 & 0.07 / 30.2 \\
Contrast ($f$) & 0.95 / 43.4 & 0.90 / 37.9 & 0.85 / 34.4 & 0.80 / 32.0 \\
\bottomrule
\end{tabular*}
\end{table}

%==========================================================
\section{Physics Simulation}
\label{sec:supp_sim}

\noindent\textbf{Setup.}
We verify the properties of delta tokens on 2D physics scenes from PhyWorld~\cite{phyworld,physar}.
Unlike real videos, simulation can re-render the same trajectory with different object appearances and provides exact ground-truth trajectories, so we can directly test whether delta tokens encode motion independently of appearance and measure how accurately the AR model predicts them.
Since these scenes contain only simple shapes on a plain background, encoders pretrained on open-domain images may encode them with priors from natural images that the scenes do not have (\eg, a red sphere on a white background might be associated with a flag).
To verify the properties of delta tokens without such confounding priors, we train the simulation tokenizer directly in pixel space.
It follows the structure of the real-world tokenizer with the differences listed in \cref{tab:tok_compare}, and its decoder also serves as the renderer, so no diffusion model is involved.
The AR model is trained for TI2V generation without V2V pretraining, where the prompt is given by the simulator parameters of the scene.

\begin{table}[t]
\centering
\caption{Settings of the simulation and real-world tokenizers.}
\label{tab:tok_compare}
\fontsize{6.5}{7.5}\selectfont\renewcommand{\arraystretch}{1.15}
\setlength{\tabcolsep}{3pt}
\begin{tabular*}{\linewidth}{@{\extracolsep{\fill}}lll@{}}
\toprule
& \textbf{Simulation} & \textbf{Real world} \\
\hline
Chunk gap & $(t, t{+}4)$ or $(t, t{+}3)$ & $(t, t{+}8)$ \\
Resolution & $256\times256$ & 480p \\
Encoder & ViT-B~\cite{vit} from scratch & Frozen SigLIP2-g \\
Encoder features & $32\times32\times768$ & $\tfrac{H}{16}\times\tfrac{W}{16}\times1536$ \\
Queries & 16 & 32 \\
FSQ levels & $(8,8)$ & $(8,5,5,5,5,5)$ \\
Decoder & ViT-B from scratch & 24 blocks from SigLIP2-g \\
Decoder usage & Renderer & Training only \\
Output & Pixels & Token distribution \\
Objective & MSE & KL \\
Trainable params & 233M & 709M \\
\bottomrule
\end{tabular*}
\end{table}

\noindent\textbf{Data.}
(1)~Training set: we take the official free-fall videos of PHYRE from 60 templates and keep 5.1M of them after removing those whose objects overlap in the first frame. We also take the official 300K videos of each of uniform motion, parabolic motion, and elastic collision, and randomly render the moving object as a square in half of them, which gives about 6.0M training videos in total (5.1M from PHYRE and 0.9M from the three single motions).
(2)~Evaluation set: for PHYRE, we apply the same filtering to the official in-template evaluation videos and keep 51,178 of them. For the single motions, we use the 379 in-distribution videos of the official evaluation set.
(3)~Preprocessing: PHYRE frames are downsampled from $512\times512$, and each video yields 12 chunks for PHYRE and 10 chunks for the single motions. During training, dynamic objects are randomly colored red, green, or blue, while static objects stay black.

\noindent\textbf{Tokenizer.}
Besides the MSE loss, it uses an FSQ entropy regularizer of weight 0.01, and it is trained for 192K steps on 7 H200 GPUs with 16 frame pairs per GPU, using a constant learning rate of $10^{-4}$ with 1,000 warmup steps and $\beta_2\!=\!0.99$.

\noindent\textbf{AR model.}
The AR model is Qwen3-VL-8B~\cite{qwen3vl} with its vision encoder replaced by the tokenizer encoder, whose first-frame features are merged in $2\times2$ groups and projected to the hidden size of the LLM by an MLP.
The input sequence consists of the scene JSON prompt, the first-frame tokens, a start marker, and the delta tokens, \ie, 192 tokens for PHYRE (12 chunks) and 160 for the single motions (10 chunks).
The JSON prompt lists the world type and the attributes of each object, such as its kind, role, position, angle, size, and speed, in a fixed key order with up to three decimals.
For example, a PHYRE scene is described as {\def\UrlBreaks{\do\,}\def\UrlBigBreaks{}\path|{"world":"phyre","objects":[{"kind":"ball","role":"dynamic","x":0.301,"y":0.712,"angle":0.0,"size":0.05},...]}|}, and a parabolic-motion scene as {\def\UrlBreaks{\do\,}\def\UrlBigBreaks{}\path|{"world":"parabola","objects":[{"kind":"ball","x":0.155,"y":0.8,"size":0.233,"speed":2.8}]}|}.
The delta tokens and 4 markers are added to the vocabulary with a new embedding table and a new output head.
All parameters except the tokenizer encoder are trained for 275K steps on 8 H200 GPUs with 8,192 tokens per GPU, using a constant learning rate of $10^{-4}$ with 500 warmup steps and $\beta_2\!=\!0.99$.
To further narrow the gap between teacher forcing and rollout, we then apply trajectory distillation (\cref{alg:td}) with the forward KL for 7K steps with 64 videos per step, using a constant learning rate of $10^{-6}$ with 50 warmup steps, in a setting adapted to simulation: $x$ contains no ground-truth tokens, and the teacher is teacher-forced on all ground-truth tokens instead of the student rollout. Starting at 0\,s, the student predicts the tokens from the first frame alone, and the teacher is teacher-forced on the whole ground truth.
The distribution of the student along its own rollout thus fits the teacher-forced distribution.
Since every position has a fixed ground-truth token, the student learns to predict the correct token even after its own errors.

\noindent\textbf{Evaluation.}
(1)~Tokenizer: we re-render each test video with the official rendering engine of PhyWorld, keeping the same trajectory but changing one appearance attribute of the objects. We re-encode the changed video and measure the fraction of delta tokens that stay unchanged. We also measure the centroid drift, \ie, the distance between the rendered and ground-truth centroids, when the decoder reconstructs the changed frame pairs with the tokens before the change (token transfer), and compare it with the lower bound given by reconstruction with their own re-encoded tokens. Radius scaling and collision are not tested together, since resizing changes the collision trajectory.
(2)~AR model: we report the token accuracy under teacher forcing and under rollout from the first frame, as well as the PSNR of the frames rendered from the predicted tokens.

\begin{table}[t]
\centering
\caption{Token invariance (\%) and centroid drift (px)$\downarrow$ in simulation for each detailed attribute. R, G, and B denote red, green, and blue, \eg, RB2RG changes the left and right balls from red and blue to red and green. For PHYRE, Color averages over random recolorings. Shape changes the ball into a square.}
\label{tab:toy_full}
\fontsize{6.5}{7.5}\selectfont\renewcommand{\arraystretch}{1.15}
\setlength{\tabcolsep}{2pt}
\begin{tabularx}{\linewidth}{@{}l|c|YY@{}}
\toprule
\multirow{2}{*}{\textbf{Attribute}} & \multirow{2}{*}{\textbf{\makecell{Token\\invariance (\%)$\uparrow$}}} & \multicolumn{2}{c}{\textbf{Centroid drift (px)$\downarrow$}} \\
& & Lower bound & Token transfer \\
\hline
\multicolumn{4}{c}{\emph{Uniform Motion}} \\
\hline
R2G & 97 & 0.36 & 0.37 \\
R2B & 95 & 0.37 & 0.37 \\
Shape & 89 & 0.37 & 0.40 \\
Radius $\times$0.8 & 87 & 0.39 & 0.40 \\
Radius $\times$0.9 & 92 & 0.37 & 0.38 \\
Radius $\times$1.1 & 93 & 0.40 & 0.43 \\
Radius $\times$1.2 & 86 & 0.46 & 0.55 \\
\hline
\multicolumn{4}{c}{\emph{Parabola}} \\
\hline
R2G & 97 & 0.40 & 0.41 \\
R2B & 96 & 0.40 & 0.41 \\
Shape & 91 & 0.51 & 0.54 \\
Radius $\times$0.8 & 86 & 0.67 & 0.67 \\
Radius $\times$0.9 & 93 & 0.47 & 0.48 \\
Radius $\times$1.1 & 92 & 0.49 & 0.52 \\
Radius $\times$1.2 & 85 & 0.78 & 0.84 \\
\hline
\multicolumn{4}{c}{\emph{Collision}} \\
\hline
RB2RG & 97 & 0.63 & 0.68 \\
RB2GR & 94 & 0.62 & 0.68 \\
RB2GB & 98 & 0.64 & 0.69 \\
RB2BR & 93 & 0.62 & 0.68 \\
RB2BG & 93 & 0.61 & 0.68 \\
Shape & 87 & 0.68 & 0.71 \\
\hline
\multicolumn{4}{c}{\emph{PHYRE}} \\
\hline
Color & 94 & 0.51 & 0.51 \\
\bottomrule
\end{tabularx}
\end{table}

\begin{table}[t]
\centering
\caption{AR prediction in simulation. TF: teacher forcing. TD: trajectory distillation.}
\label{tab:toy_ar}
\fontsize{6.5}{7.5}\selectfont\renewcommand{\arraystretch}{1.15}
\setlength{\tabcolsep}{4pt}
\begin{tabular*}{\linewidth}{@{\extracolsep{\fill}}l|*{3}{c}@{}}
\toprule
Model & TF (\%)$\uparrow$ & Rollout (\%)$\uparrow$ & PSNR (dB)$\uparrow$ \\
\hline
AR model & 89.8 & 78.1 & 36.89 \\
AR model$_{\text{w.\ TD}}$ & 89.8 & \textbf{88.4} & \textbf{38.10} \\
\bottomrule
\end{tabular*}
\end{table}

\begin{table*}[t]
\centering
\caption{Results on all dimensions of VBench-I2V. Higher is better, and the best results are in bold.}
\label{tab:vbench_i2v}
\fontsize{6.5}{7.5}\selectfont\renewcommand{\arraystretch}{1.15}
\setlength{\tabcolsep}{2pt}
\begin{tabularx}{\linewidth}{@{}l|*{12}{Y}@{}}
\toprule
\multirow{2}{*}{Model} & {\fontsize{5.5}{6.5}\selectfont Total} & {\fontsize{5.5}{6.5}\selectfont I2V} & {\fontsize{5.5}{6.5}\selectfont Quality} & {\fontsize{5.5}{6.5}\selectfont I2V} & {\fontsize{5.5}{6.5}\selectfont I2V} & {\fontsize{5.5}{6.5}\selectfont Camera} & {\fontsize{5.5}{6.5}\selectfont Subject} & {\fontsize{5.5}{6.5}\selectfont Background} & {\fontsize{5.5}{6.5}\selectfont Motion} & {\fontsize{5.5}{6.5}\selectfont Dynamic} & {\fontsize{5.5}{6.5}\selectfont Aesthetic} & {\fontsize{5.5}{6.5}\selectfont Imaging} \\
 & {\fontsize{5.5}{6.5}\selectfont Score} & {\fontsize{5.5}{6.5}\selectfont Score} & {\fontsize{5.5}{6.5}\selectfont Score} & {\fontsize{5.5}{6.5}\selectfont Subject} & {\fontsize{5.5}{6.5}\selectfont Background} & {\fontsize{5.5}{6.5}\selectfont Motion} & {\fontsize{5.5}{6.5}\selectfont Consistency} & {\fontsize{5.5}{6.5}\selectfont Consistency} & {\fontsize{5.5}{6.5}\selectfont Smoothness} & {\fontsize{5.5}{6.5}\selectfont Degree} & {\fontsize{5.5}{6.5}\selectfont Quality} & {\fontsize{5.5}{6.5}\selectfont Quality} \\
\hline
Magi-1~\cite{magi1} & \textbf{89.28} & 96.12 & \textbf{82.44} & 98.39 & 99.00 & 50.85 & 93.96 & 96.74 & 98.68 & \textbf{68.21} & 64.74 & 69.71 \\
Step-Video-TI2V~\cite{stepvideoti2v} & 88.36 & 95.50 & 81.22 & 97.86 & 98.63 & 49.23 & 96.02 & 97.06 & 99.24 & 48.78 & 62.29 & 70.44 \\
CogVideoX-I2V~\cite{cogvideox} & 86.70 & 94.79 & 78.61 & 97.19 & 96.74 & 67.68 & 94.34 & 96.42 & 98.40 & 33.17 & 61.87 & 70.01 \\
SEINE~\cite{seine} & 85.52 & 92.67 & 78.37 & 97.15 & 96.94 & 20.97 & 95.28 & 97.12 & 97.12 & 27.07 & 64.55 & 71.39 \\
I2VGen-XL~\cite{i2vgenxl} & 85.28 & 92.11 & 78.44 & 96.48 & 96.83 & 18.48 & 94.18 & 97.09 & 98.34 & 26.10 & \textbf{64.82} & 69.14 \\
ConsistI2V~\cite{consisti2v} & 84.07 & 91.91 & 76.22 & 95.82 & 95.95 & 33.92 & 95.27 & 98.28 & 97.38 & 18.62 & 59.00 & 66.92 \\
VideoCrafter-I2V~\cite{videocrafter2} & 82.57 & 86.31 & 78.84 & 91.17 & 91.31 & 33.60 & \textbf{97.86} & \textbf{98.79} & 98.00 & 22.60 & 60.78 & \textbf{71.68} \\
Cosmos-5B~\cite{cosmos} & 84.16 & 92.51 & 75.81 & 95.99 & 97.36 & 25.56 & 97.12 & 96.59 & 99.47 & 20.33 & 55.82 & 59.90 \\
Cosmos-13B~\cite{cosmos} & 84.22 & 92.60 & 75.83 & 96.17 & 97.35 & 25.43 & 97.69 & 96.77 & 99.40 & 18.70 & 55.84 & 60.15 \\
VideoMAR~\cite{videomar} & 84.82 & 94.02 & 75.61 & 97.85 & 98.38 & 21.62 & 97.13 & 97.20 & \textbf{99.57} & 10.98 & 55.81 & 62.34 \\
\hline
\ours$_{\text{w/o.\ V2V}}$ & 88.94 & 97.59 & 80.30 & \textbf{99.06} & 99.37 & 68.94 & 96.60 & 97.37 & 99.08 & 32.11 & 64.70 & 70.71 \\
\textbf{\ours (Ours)} & 89.12 & \textbf{97.80} & 80.43 & 98.98 & \textbf{99.40} & \textbf{73.92} & 96.41 & 97.59 & 98.99 & 33.74 & 64.63 & 70.95 \\
\bottomrule
\end{tabularx}
\end{table*}

\noindent\textbf{Results.}
As shown in \cref{tab:toy_full}, 93\%--98\% of the delta tokens stay unchanged when only the color changes, 87\%--91\% when the shape changes, and 85\%--93\% when the radius is scaled from 0.8 to 1.2.
It further evaluates whether the tokens encoded before the change still describe the motion of the changed video: the centroid drift with token transfer exceeds the lower bound by at most 0.09\,px in all cases.
As shown in \cref{tab:toy_ar}, trajectory distillation raises the rollout accuracy by 10.3 points to 88.4\%, close to the teacher-forcing accuracy of 89.8\%, and the PSNR of the rendered frames from 36.89 to 38.10\,dB, approaching the 40.20\,dB obtained by rendering the ground-truth tokens.

\begin{figure*}[p]
\centering
\includegraphics[width=\linewidth,height=0.86\textheight,keepaspectratio]{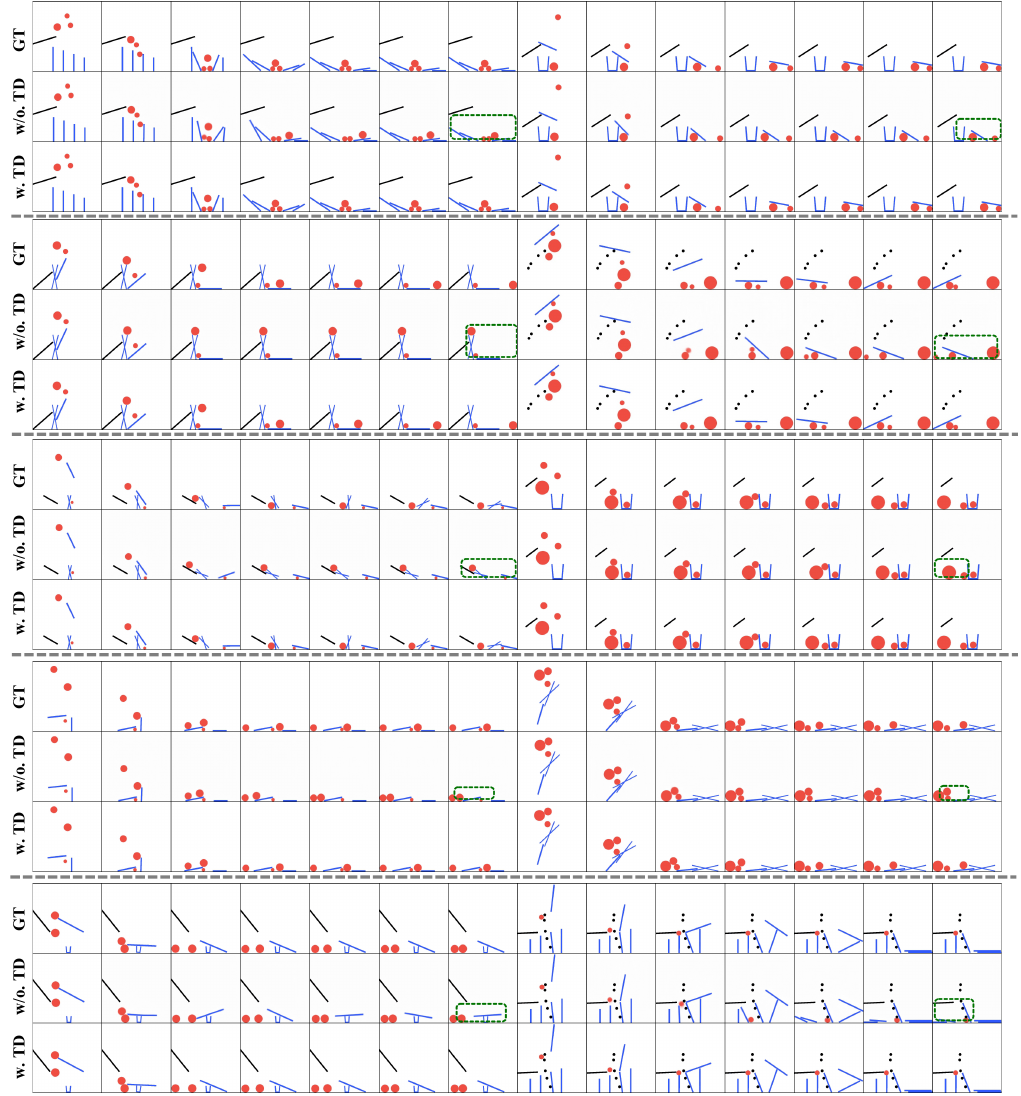}
\caption{Rollouts in simulation before and after trajectory distillation. Green boxes mark where the final configuration departs from the ground truth.}
\label{fig:supp_sim_td}
\end{figure*}

\noindent\textbf{Qualitative results.}
In \cref{fig:supp_sim_td}, we compare the rollouts of the AR model before and after trajectory distillation on PHYRE.
Before trajectory distillation, errors in early chunks accumulate, and the final configuration departs from the ground truth (green boxes).
After trajectory distillation, the AR model learns to correct such errors, and its rollouts stay close to the ground truth.

%==========================================================
\section{Additional Results}
\label{sec:supp_quant}

\subsection{Full Results on VBench-I2V}
\label{sec:supp_vbench}

\Cref{tab:vbench_i2v} reports all dimensions of VBench-I2V~\cite{vbenchpp}.
\ours$_{\text{w/o.\ V2V}}$ is trained directly for TI2V without V2V training.
\ours achieves scores comparable to those of strong video generation models on all dimensions, showing that conditioning the renderer on delta tokens preserves its ability to generate realistic videos.

\subsection{More Consistent Long Video Generation}
\label{sec:supp_long}

\begin{figure*}[p]
\centering
\includegraphics[width=\linewidth,height=0.86\textheight,keepaspectratio]{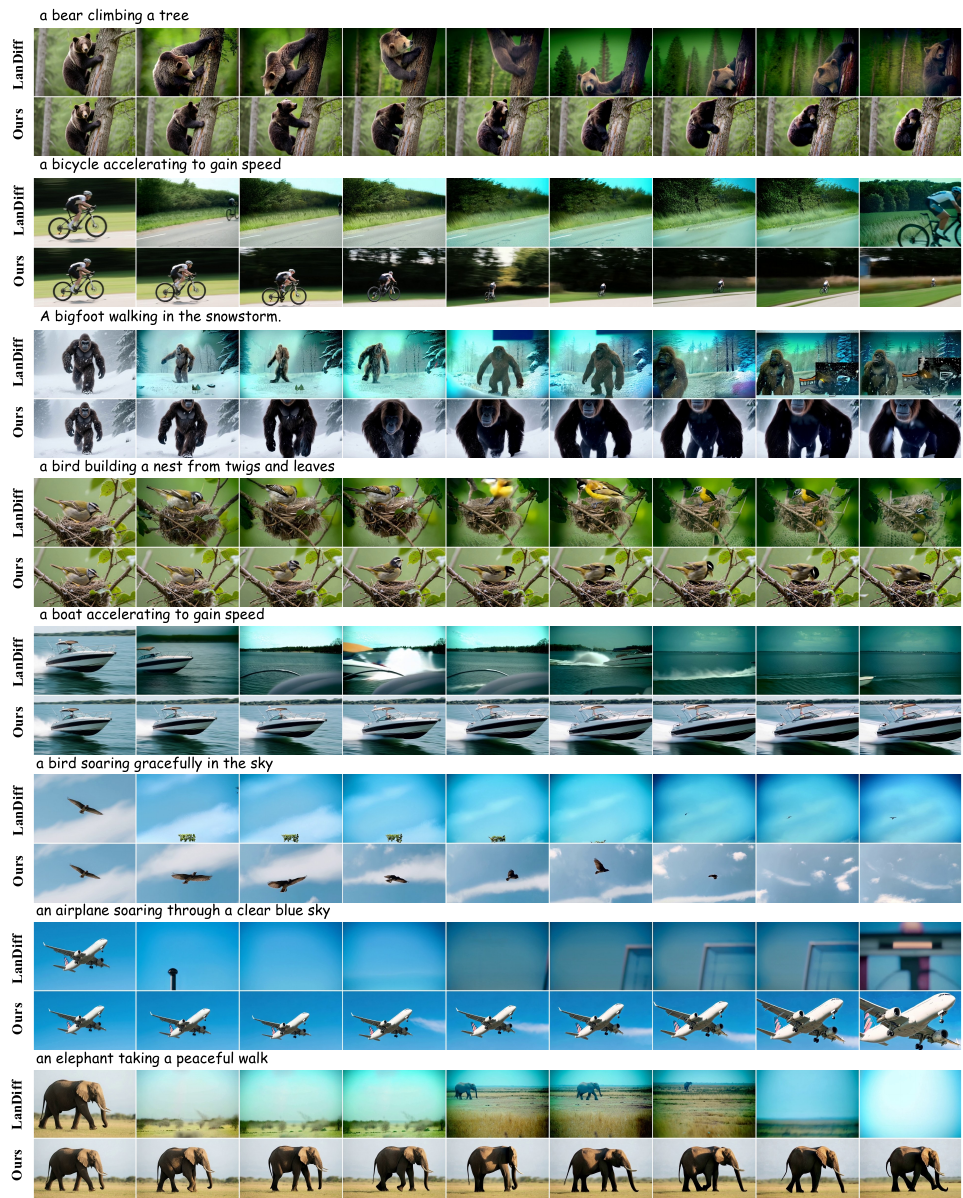}
\caption{More long videos on VBench-Long (part 1). \textit{(the supplement to \cref{fig:long})}}
\label{fig:supp_long_a}
\end{figure*}

\begin{figure*}[p]
\centering
\includegraphics[width=\linewidth,height=0.86\textheight,keepaspectratio]{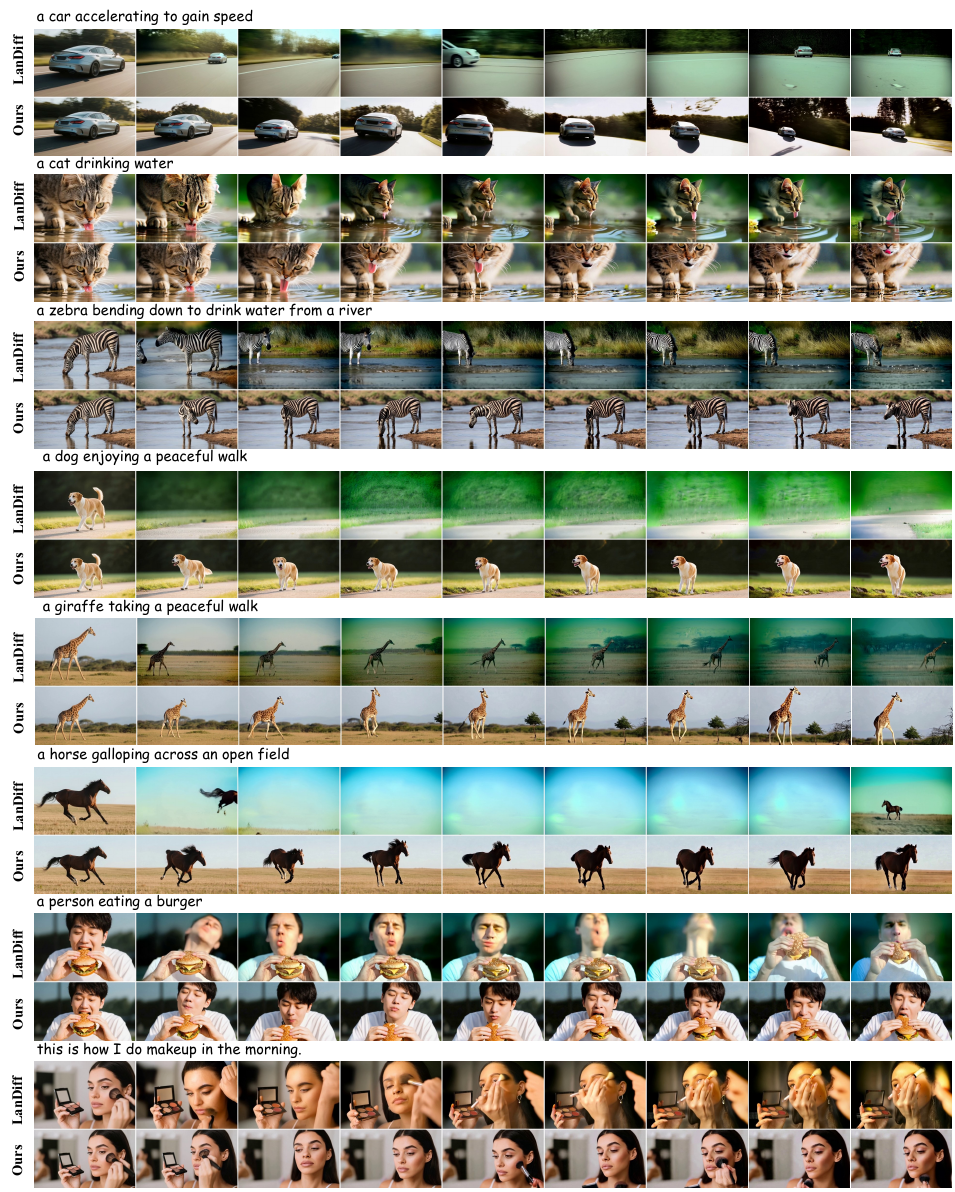}
\caption{More long videos on VBench-Long (part 2). \textit{(the supplement to \cref{fig:long})}}
\label{fig:supp_long_b}
\end{figure*}

In \cref{fig:supp_long_a,fig:supp_long_b}, we present more long videos on VBench-Long~\cite{vbenchpp}, whose first frames are generated by Wan2.2-TI2V-5B~\cite{wan22}, compared with LanDiff~\cite{landiff}.
The LanDiff videos here and in \cref{fig:long} come from its only released model, which officially generates six-second videos of 49 frames at 8\,fps and is trained on videos of at least six seconds.
Following the train-short, test-long setting of \ours, we use it to generate 25-second videos instead, by simply continuing AR sampling for a fair comparison.
The VBench-Long scores of LanDiff in \cref{tab:vbench_long} are reported in its paper.
Under the train-short, test-long setting, the 25-second videos of LanDiff gradually drift in color and content, and their subjects often blur, change identity, or disappear.
In contrast, \ours keeps the subject and the scene consistent over the whole video while the motion continues.
This benefits from the proper decoupling of reasoning and rendering, since with only 96 delta tokens per second, a 25-second video remains a sequence of 2,400 tokens that the AR model can extend chunk by chunk beyond its training horizon, while the renderer only needs to turn each segment of tokens into pixels.

\subsection{Motion Transfer across Diverse Scenarios}
\label{sec:supp_transfer}

\begin{figure*}[p]
\centering
\includegraphics[width=0.94\linewidth]{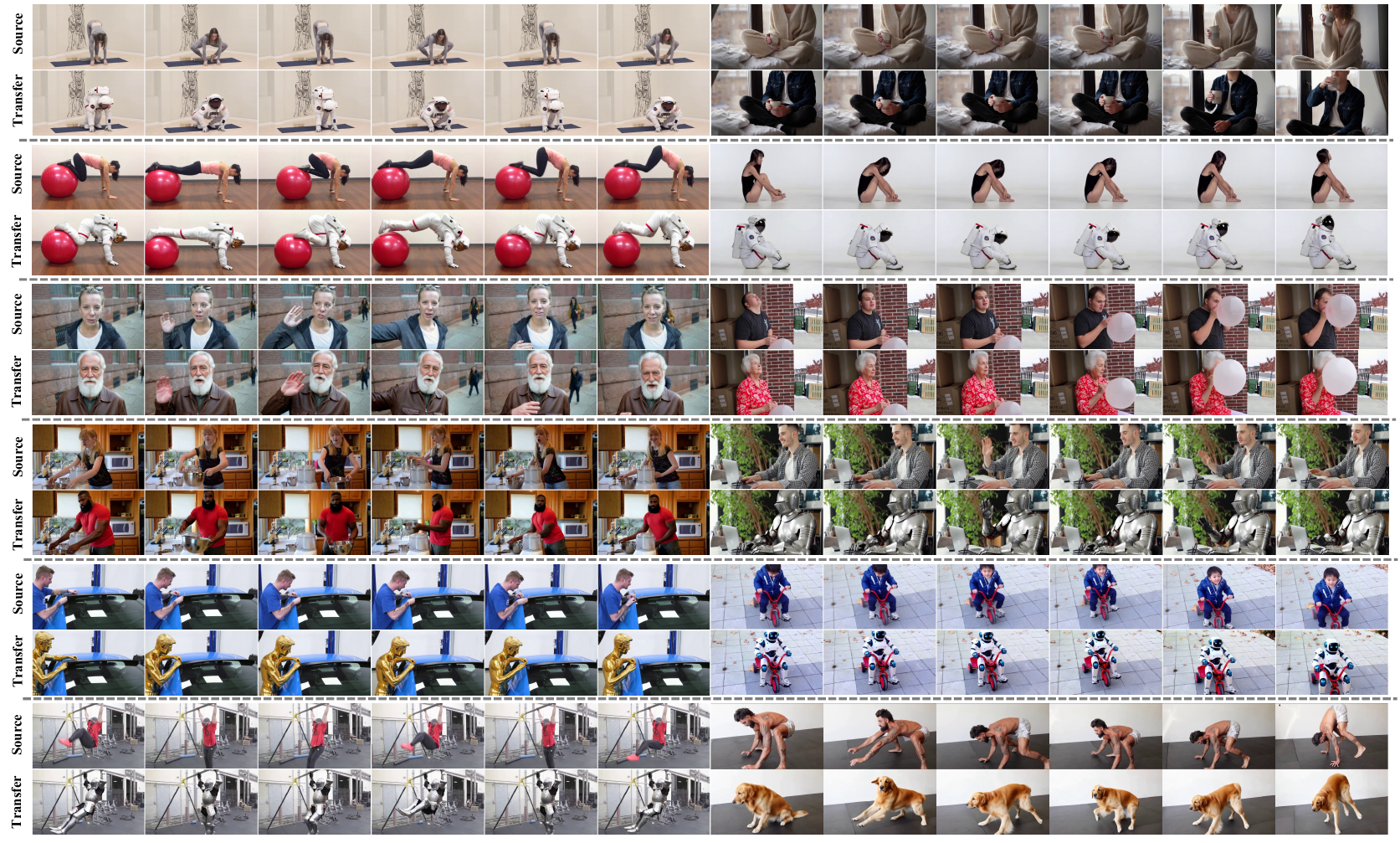}
\caption{More examples of motion transfer. \textit{(the supplement to \cref{fig:transfer})}}
\label{fig:supp_transfer}
\end{figure*}

In \cref{fig:supp_transfer}, we present more examples of motion transfer on open-domain videos.
The first frames are edited by Qwen-Image~\cite{qwenimage}, and \ours renders them with the delta tokens of the source videos without any fine-tuning.
The transferred videos follow the motion of the source videos across diverse subjects (humans, robots, an animal, and a statue) and diverse actions (full-body exercises, hand gestures, and interactions with objects).
Since the delta tokens encode how the scene evolves in a semantic space rather than the appearance of the source frames, they can drive new subjects whose appearance differs greatly from the source, and the rendered videos still largely preserve the source motion.

\subsection{More Accurate Temporal Control}
\label{sec:supp_tempo}

\begin{figure*}[p]
\centering
\includegraphics[width=0.94\linewidth]{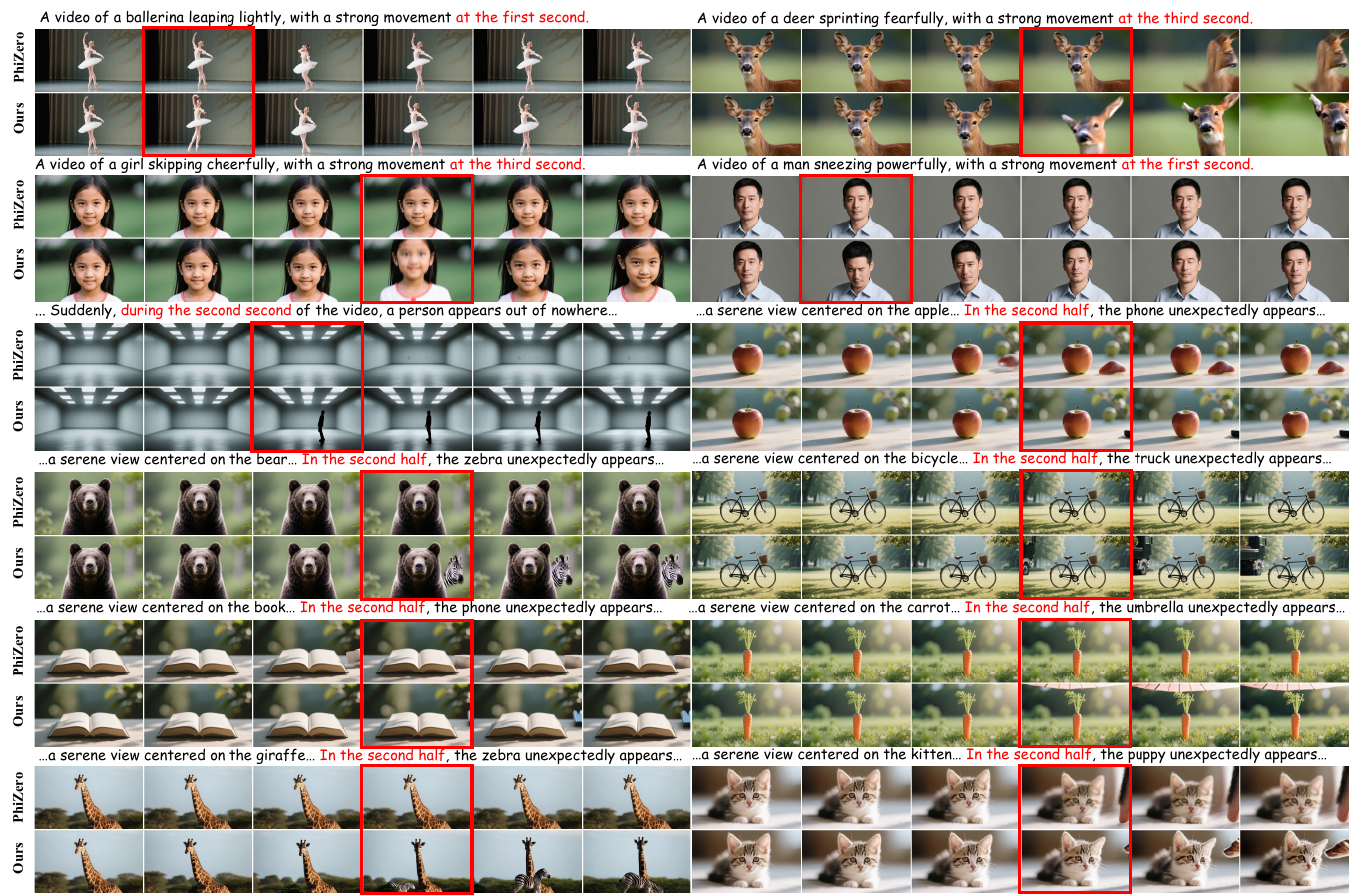}
\caption{More results of temporal control on TempoControl. \textit{(the supplement to \cref{fig:tempo_vis})}}
\label{fig:supp_tempo}
\end{figure*}

In \cref{fig:supp_tempo}, we present more results on TempoControl~\cite{tempocontrol}, compared with PhiZero~\cite{phizero}.
The first frames are generated by Wan2.2-TI2V-5B, and red boxes mark the time specified by the prompt.
With \ours, the strong movements and the newly appearing objects occur at the specified times, whereas PhiZero mostly produces nearly static videos in which these events do not happen or happen at the wrong times.
Since each chunk of delta tokens covers one third of a second, the AR model can place an event at a given second, which the clip-level tokens of PhiZero model poorly.

%==========================================================
\section{Limitations and Future Work}
\label{sec:supp_limit}

Our models are trained solely on a limited amount of open-source data, with 5.67M clips for the tokenizer and V2V pretraining and 2.03M clips for the other stages, and they face limitations in visual quality compared to baseline methods trained on significantly larger datasets. For example, Cosmos~\cite{cosmos} is trained on about 100M clips curated from 20M hours of proprietary and public videos. With such limited data, we can only fine-tune the diffusion renderer with LoRA rather than fully. The limited scale of our training data restricts the visual quality of the generated videos. 

In the future, we plan to scale up both the AR model and the renderer on larger and more diverse open-source datasets such as LAION-BVD~\cite{laionbvd}, and to fine-tune the renderer fully, so as to generate long videos of higher quality. Since the AR model and the renderer are trained in separate stages, each can be scaled independently with more data.

\end{document}